%% file: JAIR_Example_Template.tex
\documentclass[manuscript, screen]{jair}

\setcopyright{cc}
\acmDOI{10.1613/jair.1.xxxxx}

\JAIRAE{None}
\JAIRTrack{} 
\acmVolume{4}
\acmArticle{6}
\acmMonth{8}
\acmYear{2026}

\usepackage{multirow}
\usepackage{array}
\usepackage{longtable}
\usepackage{xurl}
\usepackage{placeins}
\usepackage{tikz}
\usetikzlibrary{arrows.meta,positioning,calc,fit}
\RequirePackage[
  datamodel=acmdatamodel,
  style=acmauthoryear,
  backend=biber,
  giveninits=true,
  uniquename=init
  ]{biblatex}

\begin{document}

\title{Diagnosing Evidence Utilization in Long-Context and Retrieval-Augmented Language Models under Matched Evidence Conditions}

\author{Haizhou Xia}
\email{haizhoux@outlook.com}
\affiliation{%
  \institution{University of Western Ontario}
  \city{London}
  \state{Ontario}
  \country{Canada}
}

\renewcommand{\shortauthors}{Xia}

\begin{abstract}
\textbf{Background:} Final-answer accuracy, retrieval recall, and citation overlap do not identify how much answer advantage a long-context or retrieval-augmented language model recovers from supplied evidence. A model may answer from priors, miss evidence that is present, or cite relevant text without converting it into the answer.

\textbf{Objectives:} This paper asks how to separate no-evidence answerability, oracle-evidence recoverability, full-context recovery, and retrieval-conditioned recovery under matched evaluation conditions.

\textbf{Methods:} We introduce a four-condition diagnostic protocol: no evidence, full context, retrieved evidence, and oracle-evidence reference. ONCU is used inside this protocol as a denominator-valid estimate of recovered oracle-reference evidence advantage, while answer, evidence, retrieval, robustness, and diagnostic-pattern metrics remain separate companion diagnostics. The empirical study covers five local open-weight models from the Qwen, Gemma, Llama, and Mistral families, with 18{,}000 ONCU-compatible predictions over Controlled-ONCU-safe16K, HotpotQA-ONCU, and 2WikiMultiHopQA-ONCU.

\textbf{Results:} The tested settings show a task-dependent diagnostic pattern. Controlled synthetic settings expose reduced recovery when the same evidence is embedded in long input rather than supplied compactly. In the realistic multi-hop reconstructions, full context outperforms the tested retrieved inputs in denominator-free answer/evidence metrics, with ONCU supporting the same direction on oracle-improving groups. Sensitivity audits with stronger retrieval conditions narrow some gaps but do not overturn this scoped interpretation.

\textbf{Conclusions:} The contribution is a protocol for condition-level evidence-utilization diagnosis across matched evidence roles, with denominator validity and diagnostic-pattern audits made explicit.
\end{abstract}


\maketitle

\section{Introduction}
Long-context and retrieval-augmented language models are often judged by final-answer accuracy. That score is necessary, but it does not say where the answer came from. A correct answer may come from parametric knowledge rather than the supplied context; an incorrect answer may occur even when the required passages are present; and a citation may point to useful evidence without that evidence being converted into the final answer. These cases require different diagnoses, but ordinary accuracy collapses them.

This paper studies a narrower question: when evidence is made available to a model, how much of the answer advantage made possible by that evidence is actually recovered? Answering this question requires more than a single contextual score. For the same model and examples, we need to observe what the model can answer with no evidence, what it can answer from isolated supporting evidence, and what it can answer when the evidence appears in the full context or in a retrieved subset.

We use \emph{evidence utilization} in this operational, behavioral sense. The protocol measures how observable answer and evidence scores change when evidence availability is changed while prompts, decoding, retrieval settings, and scoring rules are held fixed. It does not claim to identify the hidden computation that caused an individual answer.

The protocol has four evidence-availability conditions. The no-evidence condition estimates answerability from priors or parametric knowledge. The full-context condition tests whether evidence can be located and used when embedded in a long input. The retrieved-evidence condition tests a compact retrieve-then-read setting. The oracle-evidence reference condition tests what becomes possible when annotated supporting evidence is isolated. ONCU then normalizes a contextual condition between the no-evidence baseline and the oracle-evidence reference. Its interpretation depends on these four matched evidence roles, not on the ratio alone.

This design keeps the main diagnostic quantities separate. Answer F1 measures final answer quality but does not subtract no-evidence answerability. Evidence F1 measures overlap with supporting passages but does not show that evidence was used to form the answer. Retrieval recall measures what reaches the reader, not what the reader does with it. Oracle gaps and context gains each omit one side of the comparison. The four-condition protocol brings these terms into the same evaluation unit so that answer priors, full-context localization, retrieval-chain coverage, multi-hop integration, answer conversion, and output-format failures can be inspected separately.

The evaluation contains three ONCU-compatible benchmark components: Controlled-ONCU-safe16K, HotpotQA-ONCU, and 2WikiMultiHopQA-ONCU. Each supports no-evidence, full-context, retrieved-evidence, and oracle-evidence reference inputs for the same underlying examples. We evaluate Qwen2.5-14B, Qwen3-14B, and Gemma3-12B as the primary panel, then add Llama3.1-8B and Mistral-Small3.1-24B as a model-family extension. Retrieval-only checks, reader-facing sweeps, cross-encoder reranking, external BABILong/RULER-lite validation, and diagnostic-pattern validation are reported as sensitivity and supporting evidence rather than as separate benchmark claims.

\paragraph{Central thesis.}
The paper contributes a matched diagnostic protocol, a denominator-validity regime, and empirical diagnostic-pattern audits for behavioral evidence-utilization diagnosis. For a score field $S$ and a contextual condition $c$, the recovered evidence-advantage target $R_c=(S_c-S_{\mathrm{no}})/(S_{\mathrm{oracle}}-S_{\mathrm{no}})$ is identifiable as a condition-level diagnostic only when $S_{\mathrm{no}}$, $S_c$, $S_{\mathrm{oracle}}$, the grouping scheme, and the denominator-validity condition $S_{\mathrm{oracle}}>S_{\mathrm{no}}$ are observed under the same model and examples. This is the gap addressed by the four-condition protocol.

The paper makes five contributions:
\begin{itemize}
    \item It specifies a four-condition evidence-availability protocol that observes no-evidence, full-context, retrieved-evidence, and oracle-evidence reference behavior under the same examples, model, answer contract, retrieval controls, score field, and grouping policy.
    \item It defines a denominator-validity regime for oracle-referenced utilization: non-positive oracle-over-baseline denominators are explicitly marked, raw and clipped regimes are separated, and denominator-free answer/evidence metrics are reported alongside ONCU.
    \item It uses ONCU as the protocol-bound estimator of recovered oracle-reference evidence advantage, while making the broader contribution a matched diagnostic protocol with explicit validity checks.
    \item It gives a joint-observability argument showing why answer accuracy, evidence F1, retrieval recall, oracle gap, context gain, or a normalized contrast without matched evidence roles cannot individually identify the targeted failure modes.
    \item It provides empirical diagnostic-pattern audits across five local open-weight models and three oracle-compatible benchmark components, showing that evidence-use patterns differ between controlled and realistic settings under the tested protocol.
\end{itemize}

\section{Related Work}
This section positions the protocol relative to question answering, long-context evaluation, retrieval, grounding, and normalized performance contrasts. The key distinction is that prior evaluation families usually observe only part of the four-condition diagnostic target.

\subsection{Closed-Book, Open-Book, and No-Context Evaluation}
Closed-book question answering evaluates what a model can answer without task-specific context, while open-book and retrieval-augmented settings provide external passages before generation. This distinction is central here because no-evidence answerability is not a nuisance variable; it is one of the quantities to estimate. Standard open-book accuracy on datasets such as SQuAD or HotpotQA remains necessary, but it cannot determine whether a correct answer came from the supplied context, no-evidence priors, or parametric knowledge \parencite{rajpurkar2016squad,yang2018hotpotqa}.

\subsection{Long-Context Benchmarks and Position Sensitivity}
Long-context benchmarks test whether models can process long inputs, retrieve needles, aggregate distributed facts, or answer questions over long documents \parencite{bai2024longbench,kuratov2024babilong,hsieh2024ruler,kamradt2023needle}. Position-sensitivity work further shows that relevant evidence can be harder to use at some locations in the context \parencite{liu2024lost}. These benchmarks expose long-context failures, but they generally do not jointly estimate no-evidence answerability, full-context use, retrieved-context use, and isolated-evidence recoverability on the same examples.

\subsection{Retrieval and Grounding Evaluation}
Retrieval-augmented generation separates evidence selection from answer generation \parencite{lewis2020rag,karpukhin2020dpr}. Lexical, dense, hybrid, and rank-fusion retrieval metrics measure whether evidence is available before the reader sees the prompt \parencite{robertson2009bm25,cormack2009rrf,reimers2019sbert}. Attribution and faithfulness methods ask whether generated claims are supported by cited sources \parencite{rashkin2023ais,es2024ragas}. Both lines are complementary to the present protocol: retrieval recall is a pre-reader quantity, and citation faithfulness does not by itself show whether answer improvement came from the supplied evidence rather than a no-evidence prior.

\subsection{Normalized Gain and Oracle-Referenced Contrasts}
ONCU uses normalized-gain style reasoning, but the methodological claim is the matched evidence protocol around the ratio. Normalized contrasts and relative-improvement measures have a long history in evaluation, including information retrieval and pre/post improvement analysis \parencite{jarvelin2002cumulated,hake1998interactive}. Here, the no-evidence term estimates answer priors, the oracle-evidence term estimates an isolated-evidence reference, and the full-context or retrieved-evidence term tests whether that advantage survives the actual input regime. Without those evidence roles and the denominator-validity audit, a normalized score can inherit the ambiguities of ordinary context gain or oracle-gap reporting.

\subsection{Closest JAIR Articles and Journal-Level Positioning}
The closest JAIR articles are similar in evaluation motivation and journal-level structure rather than in proposing the same metric. \textcite{gehrmann2023cracked} argue that surface-level automatic metrics often fail to measure the intended properties of generated text. \textcite{lamalfa2024lmaas} analyze accessibility, reproducibility, reliability, and trustworthiness challenges in language-model-as-a-service systems. \textcite{gundersen2024reproducibility} articulate JAIR's reproducibility mechanisms, including structured abstracts and checklists. The present submission follows that methodological style but contributes a substantive diagnostic framework for evidence utilization in local open-weight long-context and retrieval-augmented evaluation.

\subsection{Positioning of the Present Protocol}
Table~\ref{tab:related_work_positioning} gives the one-table answer to ``why not just normalized gain or an existing retrieval/grounding metric?'' Adjacent evaluation families observe useful variables, but each omits at least one component of the full diagnostic target. The novelty is not the ratio alone. It is the joint observation of four matched evidence roles, the denominator-validity regime that marks when the oracle reference does not create recoverable advantage, and the use of companion metrics to separate reader-side recovery failures from pre-reader retrieval-chain failures.

\begin{table*}[!tbp]
\renewcommand{\arraystretch}{1.08}
\caption{Why the Four-Condition Diagnostic Protocol Is Not Reducible to a Single Prior Evaluation Line. $S_{\mathrm{no}}$ is no-evidence answerability, $S_c$ is a contextual condition score, and $S_{\mathrm{oracle}}$ is the oracle-evidence reference. The protocol is defined by joint observation under matched examples, model, score field, retriever/reader setting, grouping scheme, and denominator-validity audit.}
\label{tab:related_work_positioning}
\centering
\scriptsize
\setlength{\tabcolsep}{1.6pt}
\resizebox{\textwidth}{!}{%
\begin{tabular}{p{0.19\textwidth}ccccc p{0.28\textwidth}}
\toprule
Evaluation family & $S_{\mathrm{no}}$ & Full & Retrieved & $S_{\mathrm{oracle}}$ & Denom. audit & What remains unidentified \\
\midrule
Closed-book / no-context QA & Yes & No & No & No & No & Whether contextual evidence provides recoverable advantage. \\
Open-book / long-context QA & Usually no & Yes & Optional & Usually no & Partial & Whether the score comes from answer priors, full-context localization, or oracle-recoverable evidence use. \\
Position-sensitivity tests & Usually no & Yes & No & Usually no & No & Whether position effects reflect no-evidence priors, oracle-reference limits, or condition-specific utilization. \\
Retrieval-only evaluation & No & No & Pre-reader only & Labels optional & No & Whether retrieved evidence is converted into the final answer by the reader. \\
RAG faithfulness / attribution & Usually no & Output-level & Output-level & Evidence labels & No & Whether grounded output reflects no-evidence-adjusted recovered answer advantage. \\
Oracle-context evaluation & Usually no & Optional & Optional & Yes & Usually no & How much of the oracle-reference advantage is recovered by realistic full or retrieved inputs. \\
Context gain / oracle gap & Partial & Yes & Optional & Partial & Usually no & The missing reference side: either no-evidence answerability or recoverable oracle advantage. \\
Normalized-gain-style evaluation & Optional & Yes & Optional & Optional & Usually no & Evidence roles, retriever/reader separation, and invalid oracle-over-baseline denominators. \\
This protocol & Yes & Yes & Yes & Yes & Yes & Identifies conditional recovered oracle-reference advantage and reports companion denominator-free diagnostics. \\
\bottomrule
\end{tabular}
}
\end{table*}
\FloatBarrier

\section{Problem Formulation}

\subsection{Long-Context Question Answering}
We consider a long-context question-answering setting. Each example consists of a question $q$, a long context $C$, a gold answer $a^{\ast}$, and a set of oracle evidence passages $E^{\ast}$ when such annotations are available. The model receives the question and a condition-specific input context and produces a predicted answer $\hat{a}$ and, when required by the output contract, a set of cited evidence passages $\hat{E}$:
\begin{equation}
    x = (q, C, a^{\ast}, E^{\ast}), \qquad
    y = (\hat{a}, \hat{E}).
\end{equation}

\subsection{Evidence-Use Stages}
We define evidence utilization as the ability to transform available contextual evidence into the correct final answer. Evidence-use behavior can vary at different stages. The model may miss relevant evidence, select only a subset of the required evidence, select distractors, integrate multiple passages incompletely, or convert correctly identified evidence into the wrong answer form. The diagnostic protocol distinguishes these stages rather than collapsing them into a single final-accuracy score.

\subsection{Oracle-Reference Normalized Context Utilization}
Let $S_{\mathrm{no}}$ denote the score under the no-evidence condition, and let $S_{\mathrm{oracle}}$ denote the score under the oracle-evidence reference condition. For any evaluated contextual condition $c$, such as full context or retrieved evidence, let $S_c$ denote the corresponding score. ONCU is defined as a protocol-bound ratio:
\begin{equation}
    \mathrm{ONCU}_{\mathrm{raw}}(c) =
    \frac{S_c - S_{\mathrm{no}}}
    {S_{\mathrm{oracle}} - S_{\mathrm{no}}}.
    \label{eq:oncu}
\end{equation}
A value near 1 indicates that condition $c$ recovers most of the oracle-reference advantage. A value near 0 indicates that the condition contributes little beyond the no-evidence baseline. If $S_{\mathrm{oracle}} \leq S_{\mathrm{no}}$, ONCU is treated as invalid for that group because the oracle-evidence reference does not establish a meaningful evidence-derived advantage. For reporting aggregate results, we use a clipped version,
\begin{equation}
    \mathrm{ONCU}_{\mathrm{clip}}(c) =
    \min\left(1, \max\left(0, \mathrm{ONCU}_{\mathrm{raw}}(c)\right)\right),
    \label{eq:oncu_clipped}
\end{equation}
while retaining raw ONCU values in the released per-group outputs. Because clipping can suppress above-oracle and below-baseline behavior, the main text also includes raw-vs-clipped audit rows instead of relying only on clipped aggregates. In the experiments, ONCU is computed over metadata groups and averaged over valid groups. Consequently, ONCU should not be read as an unconditional dataset-wide average; dataset-level statements require companion answer, evidence, retrieval, and robustness metrics that do not depend on the oracle-over-baseline denominator.

\paragraph{Interpretation boundary.}
ONCU is conditional on the score field, grouping scheme, oracle-evidence construction, and denominator-valid subset. The oracle-evidence condition is an operational reference, not a perfect ceiling on attainable performance; this boundary is developed in Section~\ref{subsec:oracle_reference_not_upper_bound}. The clipped value in Eq.~\eqref{eq:oncu_clipped} supports a compact recovered-fraction summary, but raw ONCU remains the primary diagnostic audit for above-oracle behavior, below-baseline behavior, and unstable denominators. Any dataset-level conclusion therefore requires denominator-free answer/evidence metrics and retrieval diagnostics alongside ONCU.

\begin{table}[!htbp]
\renewcommand{\arraystretch}{1.10}
\caption{ONCU Interpretation Checklist. ONCU is interpreted only as a protocol-bound recovered-advantage estimate when these conditions are made explicit. The checklist separates the estimator's valid use from dataset-wide or causal claims.}
\label{tab:oncu_interpretation_checklist}
\centering
\footnotesize
\setlength{\tabcolsep}{3pt}
\begin{tabular}{p{0.30\columnwidth} p{0.33\columnwidth} p{0.29\columnwidth}}
\toprule
Requirement & Why it matters & Where this paper reports it \\
\midrule
Matched examples, model, score field, answer contract, and grouping scheme
& Prevents the ratio from comparing unmatched evidence roles.
& Sections~\ref{subsec:normalized_gain_relationship} and~\ref{subsec:companion_metrics_main}. \\

Positive oracle-over-baseline denominator
& Establishes that isolated oracle evidence provides recoverable advantage for the group.
& Eq.~\eqref{eq:oncu_validity_condition} and Table~\ref{tab:denominator_validity_audit}. \\

Raw and clipped regimes separated
& Prevents aggregate summaries from hiding above-oracle or below-baseline behavior.
& Section~\ref{subsec:raw_vs_clipped} and Appendix~\ref{app:raw_clipped_audit}. \\

Denominator-free companion metrics
& Keeps conditional ONCU from being read as an unconditional dataset score.
& Tables~\ref{tab:sci200_main_results}, \ref{tab:hotpotqa500_robustness_all}, and \ref{tab:twowiki500_main_results}. \\

Behavioral, not mechanistic, interpretation
& Avoids treating condition-level score changes as causal evidence-use attribution.
& Sections~\ref{subsec:oncu_causal_evidence_use} and~\ref{sec:limitations}. \\
\bottomrule
\end{tabular}
\end{table}

\section{Theoretical Properties of Oracle-Referenced Evaluation}
\label{sec:theoretical_properties}

This section states the formal properties needed to interpret ONCU inside the four-condition protocol. The goal is to specify when the recovered evidence-advantage contrast is identified and how its boundary cases are audited.

\subsection{From Normalized Gain to Diagnostic Estimation}
\label{subsec:normalized_gain_relationship}
Equation~\eqref{eq:oncu} becomes diagnostic because each term has a controlled evidence role. The no-evidence condition estimates what the same model can answer without contextual support; the oracle-evidence condition estimates what becomes possible when required evidence is isolated; and the full-context or retrieved-evidence condition tests whether that advantage survives the evaluated input regime. The same examples, model, score field, answer contract, decoding policy, and validity rule are held fixed.

\paragraph{Joint-observability proposition.}
Let $R_c$ denote the recovered evidence-advantage target for condition $c$. If an evaluation does not jointly observe $S_{\mathrm{no}}$, $S_c$, and $S_{\mathrm{oracle}}$ on the same examples, with the same model, score field, answer contract, and denominator-validity rule, then $R_c$ is not identified by the reported quantities. The protocol makes this target observable by fixing the three score terms and the denominator-valid subset before aggregation.

\paragraph{Proof sketch.}
Holding $S_c$ fixed while varying $S_{\mathrm{no}}$ changes the recovered fraction without changing the contextual answer score. Holding $S_c-S_{\mathrm{no}}$ fixed while varying $S_{\mathrm{oracle}}-S_{\mathrm{no}}$ changes the recovered fraction without changing context gain. Holding $S_{\mathrm{oracle}}-S_c$ fixed while varying $S_{\mathrm{no}}$ changes the recovered fraction without changing oracle gap. Retrieval recall and evidence overlap can also remain high when the reader fails to convert evidence into the answer. The detailed metric-failure table is therefore moved to Appendix~\ref{app:metric_failure_modes}.

\begin{center}
\fbox{%
\begin{minipage}{0.92\columnwidth}
\footnotesize
\textbf{Minimal counterexamples for joint observability.}
The ambiguity is visible even before the empirical results. If two systems both obtain $S_c=0.60$ but have $S_{\mathrm{no}}=0.10$ and $0.50$, the same contextual F1 reflects different evidence-derived gains. If two systems both gain $0.30$ over no evidence but have oracle recoverable advantages of $0.40$ and $0.80$, they recover different fractions of what isolated evidence makes possible. If retrieval recall is high but the reader returns the wrong answer, pre-reader evidence availability has not become answer recovery. These cases require joint observation of no-evidence, contextual, retrieved, and oracle-evidence behavior under the same examples and score field.
\end{minipage}}
\end{center}

\subsection{Why Existing Metrics Do Not Identify Evidence Utilization}
\label{subsec:metric_counterexamples}
The diagnostic target is the recovered fraction of an oracle-reference evidence advantage under a matched intervention. Full-context accuracy does not distinguish answer priors from contextual evidence use. Retrieval recall measures pre-reader availability, not reader conversion. Evidence F1 can be high when the final answer is wrong. Oracle gap ignores no-evidence answerability, while context gain ignores how much recoverable oracle-reference advantage was available. These metrics remain useful companion diagnostics, but none identifies $R_c$ alone.

\subsection{Positive Affine Invariance}
\label{subsec:affine_invariance}
Let $S'=aS+b$ for $a>0$. Then
\begin{align}
    \mathrm{ONCU}'_{\mathrm{raw}}(c)
    &= \frac{S'_c - S'_{\mathrm{no}}}{S'_{\mathrm{oracle}} - S'_{\mathrm{no}}} \\
    &= \frac{a(S_c-S_{\mathrm{no}})}{a(S_{\mathrm{oracle}}-S_{\mathrm{no}})}
     = \mathrm{ONCU}_{\mathrm{raw}}(c).
\end{align}
ONCU is therefore invariant to positive affine rescaling of the chosen score field. This property does not make all score fields interchangeable; it only means that linear rescaling of a fixed score field does not change the recovered-fraction estimate.

\subsection{Denominator Validity Condition}
\label{subsec:denominator_validity}
ONCU is meaningful only when the oracle-evidence reference provides a positive evidence-derived advantage:
\begin{equation}
    S_{\mathrm{oracle}} > S_{\mathrm{no}}.
    \label{eq:oncu_validity_condition}
\end{equation}
If the denominator is zero or negative, the ratio no longer measures recovery of an available evidence advantage. Such cases can occur when the model already answers well without evidence, when the oracle snippet is too narrow, when neighboring context helps, or when the score field is insensitive to the added evidence. The experiments therefore report valid-group counts before ONCU is interpreted and pair ONCU with denominator-free answer/evidence metrics.

\subsection{Oracle Reference and Raw-vs-Clipped ONCU}
\label{subsec:oracle_reference_not_upper_bound}
\label{subsec:raw_vs_clipped}
The oracle-evidence condition is an isolated-evidence reference, not a guaranteed ceiling. Annotated evidence may omit neighboring sentences, aliases, or redundant support that full or retrieved contexts still contain. Raw ONCU can therefore exceed 1 or fall below 0. Values above 1 can indicate useful auxiliary context or oracle-snippet incompleteness; values below 0 can indicate context-induced degradation. The clipped value in Eq.~\eqref{eq:oncu_clipped} is used for readable aggregate summaries, while raw values, valid indicators, and denominators remain available in the released per-group files.

\subsection{ONCU as a Within-Model Diagnostic Ratio}
\label{subsec:not_model_ranking}
ONCU is a within-model, within-task diagnostic ratio. It is interpreted with the underlying no-evidence, contextual, oracle-evidence, answer, and evidence scores. The goal is to separate the observed roles of long-context localization, retrieval coverage, multi-hop integration, answer conversion, and denominator validity within a matched evaluation unit.

\subsection{ONCU and Causal Evidence Use}
\label{subsec:oncu_causal_evidence_use}
ONCU is a behavioral, condition-level estimate. It measures how answer and evidence scores change when observable evidence availability is controlled under a fixed prompt, retrieval, decoding, and scoring pipeline. The resulting analysis supports condition-level contrasts over tested inputs and observed score changes; mechanistic claims about internal computations require separate causal interventions.

\section{Benchmark Construction}

\subsection{Overview}
We construct a diagnostic evaluation suite for long-context evidence utilization. The reported ONCU evaluation uses three oracle-compatible benchmark components: a controlled synthetic component, a HotpotQA-derived multi-hop component, and a 2WikiMultiHopQA-derived multi-hop component. These components include evidence annotations that allow the same example to be evaluated under no-evidence, full-context, retrieved-evidence, and oracle-evidence conditions. We additionally include BABILong-200 and RULER-lite-240 as external answer-performance validation settings. These external settings are useful for testing reasoning-in-a-haystack and synthetic long-context behavior, respectively, but they are not used for ONCU computation in this study because the current adapters do not provide the full oracle-reference condition set in the same format as the oracle-compatible components.

\begin{figure}[!htbp]
\centering
\definecolor{oncuInk}{HTML}{2F3A45}
\definecolor{oncuLine}{HTML}{6B7280}
\definecolor{oncuPanel}{HTML}{F7F8FA}
\definecolor{oncuBlue}{HTML}{E7F0FA}
\definecolor{oncuBlueDark}{HTML}{315D8C}
\definecolor{oncuGreen}{HTML}{E8F4EE}
\definecolor{oncuGreenDark}{HTML}{2F6B4F}
\definecolor{oncuGold}{HTML}{F8F1DE}
\definecolor{oncuGoldDark}{HTML}{7A5A16}
\resizebox{0.97\textwidth}{!}{%
\begin{tikzpicture}[
  font=\footnotesize,
  >=Latex,
  stage/.style={
    draw=oncuLine,
    fill=oncuPanel,
    rounded corners=2pt,
    line width=0.45pt,
    align=center,
    inner xsep=7pt,
    inner ysep=5pt,
    minimum height=10mm
  },
  cond/.style={
    draw=oncuBlueDark,
    fill=oncuBlue,
    rounded corners=2pt,
    line width=0.55pt,
    align=center,
    inner xsep=5pt,
    inner ysep=5pt,
    minimum width=0.205\textwidth,
    minimum height=17mm
  },
  audit/.style={
    draw=oncuGoldDark,
    fill=oncuGold,
    rounded corners=2pt,
    line width=0.55pt,
    align=center,
    inner xsep=7pt,
    inner ysep=5pt,
    minimum width=0.55\textwidth,
    minimum height=12mm
  },
  output/.style={
    draw=oncuGreenDark,
    fill=oncuGreen,
    rounded corners=2pt,
    line width=0.55pt,
    align=center,
    inner xsep=7pt,
    inner ysep=5pt,
    minimum width=0.78\textwidth,
    minimum height=13mm
  },
  arrow/.style={->, line width=0.55pt, draw=oncuLine}
]
\node[stage, minimum width=0.82\textwidth] (unit)
  {\textbf{Matched evaluation unit}\\
  Same examples, model, prompts, answer contract, score field, and grouping scheme};

\node[cond, below=8mm of unit, xshift=-0.3225\textwidth] (no)
  {\textbf{No evidence}\\[2pt] question only\\[-1pt] $S_{\mathrm{no}}$};
\node[cond, right=4mm of no] (full)
  {\textbf{Full context}\\[2pt] complete long input\\[-1pt] $S_{\mathrm{full}}$};
\node[cond, right=4mm of full] (ret)
  {\textbf{Retrieved evidence}\\[2pt] compact retrieved input\\[-1pt] $S_{\mathrm{ret}}$};
\node[cond, right=4mm of ret] (oracle)
  {\textbf{Oracle-evidence reference}\\[2pt] isolated supporting evidence\\[-1pt] $S_{\mathrm{oracle}}$};

\node[audit, below=10mm of $(full)!0.5!(ret)$] (valid)
  {\textbf{Denominator-valid audit}\\[-1pt]
  interpret ONCU only when $S_{\mathrm{oracle}}>S_{\mathrm{no}}$};

\node[output, below=7mm of valid] (diag)
  {\textbf{Protocol-bound outputs}\\[-1pt]
  recovered oracle-reference advantage (ONCU) + denominator-free answer/evidence metrics, retrieval diagnostics, and diagnostic summaries};

\draw[arrow] (unit.south) -- ++(0,-3mm) -| (no.north);
\draw[arrow] (unit.south) -- ++(0,-3mm) -| (full.north);
\draw[arrow] (unit.south) -- ++(0,-3mm) -| (ret.north);
\draw[arrow] (unit.south) -- ++(0,-3mm) -| (oracle.north);
\draw[arrow] (no.south) |- (valid.west);
\draw[arrow] (full.south) -- ++(0,-4mm) -| (valid.north);
\draw[arrow] (ret.south) -- ++(0,-4mm) -| (valid.north);
\draw[arrow] (oracle.south) |- (valid.east);
\draw[arrow] (valid.south) -- (diag.north);

\node[draw=oncuLine, rounded corners=3pt, fit=(unit)(no)(oracle)(diag),
      inner sep=6pt, line width=0.35pt] {};
\end{tikzpicture}
}
\caption{Matched four-condition diagnostic protocol. The four evidence-availability conditions are observed on the same examples under one scoring contract. ONCU is computed only after the denominator-validity audit and is interpreted with denominator-free answer/evidence metrics, retrieval diagnostics, and diagnostic summaries.}
\Description{
A compact protocol diagram. The same examples, model, prompts, answer contract,
score field, and grouping scheme are evaluated under no-evidence, full-context,
retrieved-evidence, and oracle-evidence reference conditions. A denominator-validity
audit precedes ONCU and companion diagnostics.
}
\label{fig:oncu_workflow}
\end{figure}
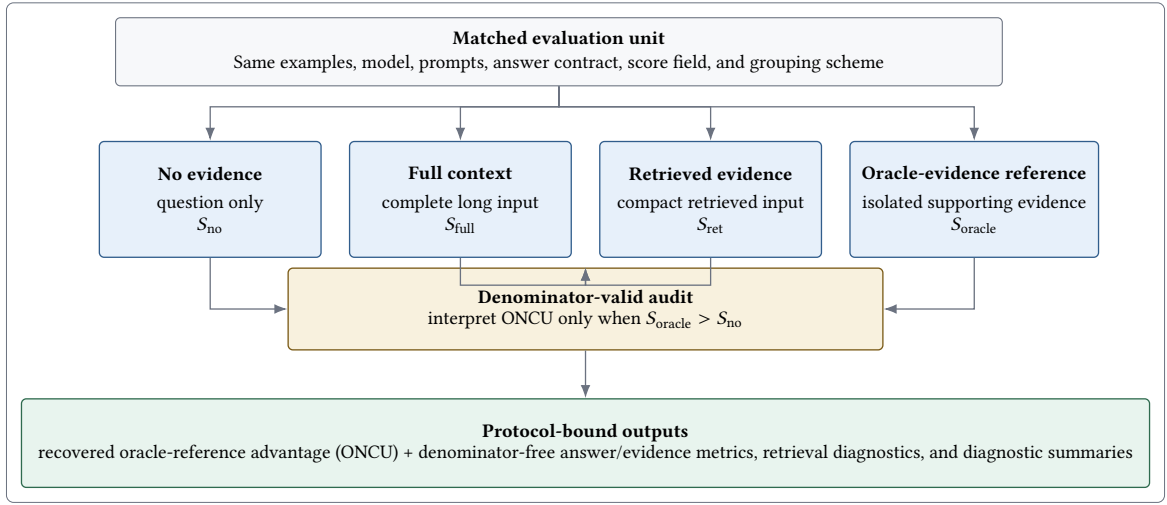

Figure~\ref{fig:oncu_workflow} summarizes the core diagnostic protocol. External BABILong-200 and RULER-lite-240 checks are reported separately as answer-performance validation because they do not instantiate the complete oracle-reference protocol.

\begin{table}[!htbp]
\renewcommand{\arraystretch}{1.12}
\caption{Benchmark Components and Their Roles in the Diagnostic Framework.}
\label{tab:benchmark_overview}
\centering
\footnotesize
\setlength{\tabcolsep}{2.5pt}
\begin{tabular}{p{0.30\columnwidth} p{0.36\columnwidth} c c}
\toprule
Component & Role & Oracle Ev. & ONCU \\
\midrule
Controlled-ONCU
& Main controlled evaluation
& Yes
& Yes \\

Controlled-ONCU-safe16K
& Size-comparable model comparison
& Yes
& Yes \\

HotpotQA-ONCU
& Realistic multi-hop evaluation
& Yes
& Yes \\

2WikiMultiHopQA-ONCU-500
& Additional realistic multi-hop validation
& Yes
& Yes \\

BABILong-200
& External answer validation
& No
& No \\

RULER-lite-240
& External synthetic long-context validation
& No
& No \\
\bottomrule
\end{tabular}
\end{table}

\begin{table*}[!tbp]
\renewcommand{\arraystretch}{1.10}
\caption{Scope and Possible Selection Bias of the Oracle-Compatible Components. The ONCU-compatible datasets are constructed to expose the four evidence conditions, not to preserve every property of the original benchmarks.}
\label{tab:dataset_scope_bias}
\centering
\scriptsize
\setlength{\tabcolsep}{2pt}
\begin{tabular}{p{0.18\textwidth} p{0.15\textwidth} p{0.23\textwidth} p{0.13\textwidth} p{0.20\textwidth}}
\toprule
Component & Source & Inclusion rule & Evaluated sample size & Possible bias or boundary \\
\midrule
Controlled-ONCU-safe16K
& Synthetic generator
& Generated examples with known evidence, metadata cells, and safe context lengths.
& 200 core examples; 3,200 scaling examples.
& Clean evidence structure supports mechanism diagnosis but may underrepresent natural ambiguity. \\

HotpotQA-ONCU
& HotpotQA
& Retained examples whose supporting facts can be reliably aligned to visible passage identifiers.
& 200 core examples; 500 robustness examples.
& May favor cleaner evidence alignments and should not be read as an unconditional HotpotQA benchmark. \\

2WikiMultiHopQA-ONCU
& 2WikiMultiHopQA
& Retained examples with evidence paths that can be converted into oracle passages.
& 500 validation examples.
& May underrepresent examples with ambiguous, redundant, or hard-to-map evidence paths. \\

BABILong-200 and RULER-lite-240
& External long-context benchmarks
& Adapted for answer-performance validation only.
& 200 and 240 examples.
& Current adapters do not instantiate the full no-evidence/oracle-evidence protocol needed for ONCU. \\
\bottomrule
\end{tabular}
\end{table*}

\subsection{Sample Selection and Stratification}
All reported datasets are materialized as fixed processed JSONL files before model evaluation and are referenced by configuration files during inference. Sample inclusion is determined only by the dataset-construction pipeline and metadata filters, not by model outputs. This prevents post-hoc selection of examples after observing model behavior.

Table~\ref{tab:dataset_scope_bias} makes the construction boundary explicit. The controlled component is designed for diagnostic control rather than naturalistic coverage. The HotpotQA and 2WikiMultiHopQA components are reconstructed into oracle-compatible, passage-identified versions, so their conclusions apply to those retained evaluation sets. The external BABILong and RULER-lite checks are included only as answer-performance validations because their current adapters do not provide the complete four-condition reference structure.

For the controlled benchmark, the full generator supports 4K, 8K, 16K, and 32K context settings. The primary cross-model comparison uses the Controlled-ONCU-safe16K-200 subset, which is drawn from the 4K, 8K, and 16K controlled settings after excluding 32K contexts. The exclusion is applied before evaluation to avoid backend-specific truncation near the maximum context window when task instructions, question text, passage identifiers, and structured-output constraints are included. Each controlled sample retains metadata for context length, evidence position, distractor similarity, and reasoning type. The subset is selected to cover the available metadata cells in the safe 16K range, and the ONCU aggregation is later performed over valid metadata groups rather than directly over examples so that larger cells do not dominate the diagnostic summary.

For HotpotQA-ONCU, examples are first converted into passage-identified long-context instances. Supporting facts are aligned to visible passages, and an example is retained only when the required supporting facts can be reliably mapped to oracle evidence passage identifiers. The gold evidence identifiers and distractor labels are stored in metadata and are not exposed to the model. The HotpotQA-ONCU-200 set is used in the balanced 3-model by 2-dataset main matrix. The HotpotQA-500 robustness set is constructed with the same filtering rules, the same passage-identifier format, the same top-$k=3$ main retrieval protocol, and a fixed random seed of 42. The 500-sample setting therefore changes sample size, not the task definition or evaluation contract.

Across both benchmark components, the no-evidence, full-context, retrieved-evidence, and oracle-evidence conditions are produced from the same underlying examples. Within a model--dataset run, condition differences therefore reflect evidence availability rather than changes in the evaluated sample pool.

For 2WikiMultiHopQA-ONCU-500, examples are converted from the 2WikiMultiHopQA multi-hop question-answering dataset, which provides evidence information containing reasoning paths for multi-hop questions \parencite{ho2020wikimultihopqa}. We construct 500 passage-identified long-context instances using the same four-condition diagnostic protocol as the other ONCU-compatible components. Each retained example contains a gold answer, oracle evidence passages, a full passage-annotated context, and distractor passages. The resulting 500-sample set is used as an additional realistic multi-hop validation component rather than as a replacement for the balanced 200-sample core matrix.

For BABILong-200, we construct an external validation set from four context configurations, 0k, 1k, 2k, and 4k, and five task types, qa1, qa2, qa3, qa6, and qa7, using 10 examples per task--configuration cell. This yields 200 examples before model evaluation. BABILong samples are evaluated under no-evidence, full-context, and retrieved-evidence conditions using the same deterministic decoding and lexical retrieval settings as the main experiments. Because the current BABILong adapter does not provide oracle-evidence annotations compatible with ONCU, no oracle-evidence condition is constructed and BABILong is reported only as external answer-performance validation.

For RULER-lite-240, we construct an external synthetic long-context validation set with three task families, \texttt{retrieval\_key\_value}, \texttt{multi\_hop\_trace}, and \texttt{aggregation\_sum}, four context lengths, 4K, 8K, 16K, and 32K, and 20 examples per task--length cell. This yields 240 examples before model evaluation. RULER-lite samples are evaluated under full-context and retrieved-context conditions with the same three evaluated models and a fixed top-$k=3$ lexical retrieval setting. Because this adapter does not instantiate no-evidence and oracle-evidence references compatible with ONCU, RULER-lite is reported only as external answer-performance validation.

\subsection{Controlled-ONCU}
The controlled component aims to isolate the factors affecting long-context evidence utilization. It systematically varies four dimensions: context length, evidence position, distractor similarity, and reasoning type.
\begin{itemize}
    \item \textbf{Context length:} 4K, 8K, 16K, and 32K tokens.
    \item \textbf{Evidence position:} front, middle, end, and scattered in the main controlled generator; decile midpoint positions in the length--position scaling extension.
    \item \textbf{Distractor similarity:} none, low, high, and conflicting.
    \item \textbf{Reasoning type:} single-hop, multi-hop, comparison, and arithmetic.
\end{itemize}
This design enables controlled analysis of where and why long-context models fail.

The main cross-model controlled comparison uses the safe16K subset described above. To test whether the aggregate controlled full-context pattern hides systematic scaling behavior, we additionally construct a controlled length--position scaling extension. This extension crosses four context lengths, 4K, 8K, 16K, and 32K, with ten decile evidence-position buckets, \texttt{pos\_00} through \texttt{pos\_09}, four distractor-similarity settings, four reasoning types, and five random seeds per cell. The resulting processed input contains 3,200 samples. It is used as a diagnostic extension rather than as a replacement for the balanced three-model safe16K comparison.

\subsection{HotpotQA-ONCU}
The HotpotQA-derived component provides a realistic multi-hop question-answering setting \parencite{yang2018hotpotqa}. We align supporting facts to passages and retain only examples whose supporting facts can be reliably mapped to oracle evidence passages. These samples allow ONCU computation in a real multi-hop setting.

\subsection{2WikiMultiHopQA-ONCU}
The 2WikiMultiHopQA-derived component provides a second realistic multi-hop setting. 2WikiMultiHopQA was designed to evaluate reasoning steps by combining structured and unstructured information and by providing evidence information containing reasoning paths for multi-hop questions \parencite{ho2020wikimultihopqa}. This makes it well aligned with oracle-referenced evaluation: the oracle-evidence reference can be constructed from the annotated evidence path, while the full-context and retrieved-evidence conditions test whether the model or retriever preserves and uses the required evidence chain. We report a 500-sample 2WikiMultiHopQA-ONCU validation set for all three evaluated models.

\subsection{External Long-Context Validation}
LongBench, BABILong, and RULER are important long-context benchmarks for evaluating model robustness beyond a single controlled task family \parencite{bai2024longbench,kuratov2024babilong,hsieh2024ruler}. In this study, we report BABILong-200 as an external reasoning-in-a-haystack validation setting and RULER-lite-240 as an external synthetic long-context validation setting. LongBench is treated as future external validation because its heterogeneous task types require additional adapter design to define comparable evidence references and metrics. The current BABILong and RULER-lite adapters do not instantiate the same oracle-referenced four-condition protocol as Controlled-ONCU, HotpotQA-ONCU, and 2WikiMultiHopQA-ONCU; therefore, both external settings are interpreted separately from the oracle-referenced core results.

\subsection{Passage Annotation and Leakage Prevention}
All visible passages are assigned neutral passage identifiers such as \texttt{[passage\_id: p0001]}. Gold evidence and distractor labels are stored only in metadata and are not exposed to the model. This prevents label leakage and ensures that models must infer relevance from passage content rather than explicit evidence markers.

\section{Diagnostic Protocol}

\subsection{Core Diagnostic Conditions}
The main evaluation uses fixed diagnostic conditions rather than prompt optimization. Each condition changes the evidence available to the model while keeping the answer contract, decoding policy, and evaluation metrics fixed.

\subsubsection{No-Evidence Condition}
The no-evidence condition provides only the question. It estimates answer priors and parametric knowledge. This condition is needed for realistic question-answering datasets, where a model may answer some questions without using the supplied context.

\subsubsection{Full-Context Condition}
The full-context condition provides the complete long context and asks the model to answer while citing supporting passage identifiers. This is the primary condition for measuring whether evidence embedded in a long input is used effectively.

\subsubsection{Retrieved-Evidence Condition}
The retrieved-evidence condition applies a deterministic lexical retriever to select a compact evidence set before answer generation, following the role of lexical retrieval baselines such as BM25 in information retrieval \parencite{robertson2009bm25}. In the main diagnostic matrix, retrieval is configured with top-$k=3$, chunk size 220, and overlap 40. Candidate chunks are produced from the same passage-annotated context used in the full-context condition, and retrieval is applied before answer generation using the question as the query. The selected chunks are then passed to the same answer-generation contract used by the other contextual conditions. This matched diagnostic condition asks whether a fixed compact retrieved context recovers the evidence-derived advantage observed under the no-evidence, full-context, and oracle-evidence references. In the retrieval-budget ablation, only top-$k$ is varied, while chunk size, overlap, decoding policy, output contract, and evaluation pipeline are held fixed.

Retrieval family and retrieval depth are treated as explicit protocol variables rather than hidden implementation details. We therefore run three retriever-family checks. The first is a retrieval-only audit that holds the processed examples, passage segmentation, chunk size, overlap, and evidence-overlap metrics fixed while varying the retrieval family: lexical retrieval, off-the-shelf dense sentence-embedding retrieval, hybrid lexical--dense rank fusion using reciprocal ranks, a deterministic iterative query-expansion baseline, and an oracle retrieval reference. This audit diagnoses evidence-chain availability, ranking, and distractor exposure before downstream answer generation. The second is a matched retriever-family ONCU sensitivity experiment that reruns the complete no-evidence, full-context, retrieved-evidence, and oracle-evidence protocol for dense@16 and hybrid@16 retrieved inputs on HotpotQA-ONCU-200 and 2WikiMultiHopQA-ONCU-500. The third is a broader reader-facing validation that reruns answer generation for lexical, dense, and hybrid retrieved contexts at top-$k \in \{3,8,16\}$ and reports answer/evidence metrics across budgets. The matched sensitivity experiment tests whether the denominator-valid ONCU contrast and denominator-free answer/evidence contrast remain stable under stronger retrieved inputs; the broader reader-facing validation tests whether retrieval-family improvements transfer to downstream answer generation.

\subsubsection{Oracle-Evidence Condition}
The oracle-evidence condition provides only the gold supporting evidence. It serves as an oracle-evidence reference for ONCU normalization and verifies whether the model can answer when the necessary evidence is isolated. Because retrieved chunks may contain oracle passages plus adjacent local context, the oracle-evidence condition is an empirical reference instead of a guaranteed upper bound for every individual group.

\subsection{Auxiliary Diagnostic Probes}
Auxiliary probes are used only for failure analysis. The concise-reasoning probe tests whether a fixed reasoning-style instruction changes utilization behavior. The evidence-selection probe separates evidence selection from answer generation. The evidence-sufficiency probe adds a verification and expansion step to reveal missing-support failures. These probes are not optimized per model or per dataset and are not used as the basis for the main ONCU claims.

\subsection{Fixed Protocol and Reproducibility}
All instruction templates are fixed before model comparison and shared across models and datasets. We do not tune templates for individual models, datasets, or failed examples. Local Ollama experiments use deterministic decoding with temperature set to 0 and an explicit context-window configuration of 32,768 tokens. The retrieved-evidence condition uses a fixed deterministic lexical retrieval configuration. Each run records configuration files, per-sample metrics, ONCU summaries, and protocol metadata for reproducibility.

\paragraph{Runtime and model-environment record.}
The released artifact includes \path{RUNTIME_REPRODUCIBILITY_RECORD.md} and \path{runtime_reproducibility_record.json}, generated by \path{scripts/export_runtime_record.py}. The record captures the available runtime metadata for the reproduction host, including operating system, Python executable, package versions, PyTorch/CUDA status, GPU name and memory, \texttt{nvidia-smi} output, deterministic inference controls, model tags requested by the protocol, context-window settings, and any unavailable runtime commands. The submitted Runpod record reports Linux/x86\_64, Python~3.11.10, PyTorch~2.4.1+cu124, an NVIDIA~A40 GPU with 44.43~GB reported memory, NVIDIA driver~570.211.01, and CUDA~12.8. Model identifiers, retrieval settings, decoding controls, context length, and output paths are also fixed in the released configuration files and reproduction README.

\paragraph{Hyperparameter-search boundary.}
The experiments are diagnostic protocol runs rather than open-ended hyperparameter optimization. Final reported settings are fixed in \path{configs/}; the explored sensitivity dimensions are explicitly enumerated in the manuscript and artifacts: retrieval budgets, lexical/dense/hybrid retriever families, dense@16 and hybrid@16 matched ONCU sensitivity, cross-encoder reranking candidate/final budgets, controlled context length and evidence position, model-family extension, and external BABILong/RULER-lite validation settings. No additional unpublished prompt tuning, model-specific decoding search, or per-dataset failed-example tuning is used to select the main reported results.

\section{Evaluation Metrics}

\subsection{Answer Metrics}
We report both strict and relaxed answer metrics. Strict exact match and strict F1 require the predicted answer to match the gold answer precisely. Relaxed metrics apply normalization such as lowercasing, accent removal, punctuation removal, and synthetic suffix removal when appropriate.

\subsection{Evidence Metrics}
We evaluate evidence selection using evidence precision, recall, and F1:
\begin{equation}
    P_E = \frac{|\hat{E} \cap E^{\ast}|}{|\hat{E}|},
\end{equation}
\begin{equation}
    R_E = \frac{|\hat{E} \cap E^{\ast}|}{|E^{\ast}|},
\end{equation}
\begin{equation}
    F1_E = \frac{2P_E R_E}{P_E + R_E}.
\end{equation}

\subsection{ONCU Metrics}
We compute ONCU using multiple score fields, including strict answer F1, relaxed answer F1, strict exact match, and relaxed exact match. Unless otherwise specified, ONCU-Relaxed-F1 refers to ONCU computed with relaxed answer F1 as $S$ in Eq.~\eqref{eq:oncu}. This is the primary ONCU metric because it accounts for semantically correct answers that differ only in formatting or synthetic identifiers. The alternative strict-F1 and exact-match variants are treated as score-field sensitivity checks; the main claims are stated only where ONCU is interpreted together with denominator-free answer/evidence metrics and denominator-validity audits.

\subsection{Companion Metrics and Audit Outputs}
\label{subsec:companion_metrics_main}
ONCU is interpreted together with denominator-free answer and evidence metrics, retrieval diagnostics, bootstrap intervals, paired contrasts, and diagnostic-pattern audits. The main text uses those quantities only where they support the central diagnostic argument; the complete alternative-metric tables and representative case studies are moved to Appendix~\ref{app:metric_comparison} to keep the exposition focused.
\subsection{Bootstrap Confidence Intervals}
To assess statistical reliability, we compute non-parametric bootstrap confidence intervals for the final 200-sample matrix, the HotpotQA-500 robustness checks, and the BABILong-200 external validation. For answer and evidence metrics, examples are resampled with replacement within each model--dataset--condition group. For ONCU, valid metadata groups are resampled with replacement within each model--dataset--condition group, because ONCU is computed as a group-normalized quantity. The final 200-sample analysis, the HotpotQA-500 robustness analysis, and the BABILong-200 answer-performance analysis use 5,000 bootstrap replicates, and intervals are reported as two-sided 95\% percentile intervals. These intervals are used only as reliability checks for the diagnostic trends and are not used to tune prompts, retrieval settings, or model-specific decoding policies.

\subsection{Paired Effect Sizes and Multiple-Comparison Control}
The core diagnostic protocol is a repeated-measures design: the same examples are evaluated under no-evidence, full-context, retrieved-evidence, and oracle-evidence conditions. We therefore report paired condition contrasts in addition to aggregate means. For a contrast such as retrieved evidence minus full context, we align examples by sample identifier, compute the per-sample score difference, and report the paired mean difference, a paired bootstrap 95\% confidence interval, and a standardized paired effect size. The standardized effect size is the mean paired difference divided by the standard deviation of the paired differences. These quantities are intended to measure effect magnitude, not merely statistical detectability.

For the controlled length--position scaling analysis, the same generated cells are evaluated across context lengths and evidence-position buckets. We therefore compute paired length and position contrasts over matched position or length--position cells, and we supplement them with regression-style diagnostics over aggregated ONCU cells. For the retrieval-family ablation, we compute paired contrasts over sample identifiers because each retriever and top-$k$ setting is evaluated on the same underlying examples.

When multiple confirmatory contrasts are reported within the same analysis family, we apply Holm adjustment to the normal-approximation diagnostic $p$-values \parencite{holm1979sequential}. For exploratory regression-style diagnostics, we additionally report Benjamini--Hochberg false-discovery-rate adjusted values \parencite{benjamini1995fdr}. The interpretation of the results emphasizes effect sizes and confidence intervals rather than isolated $p$-values, following the general caution that statistical significance alone does not measure effect size or scientific importance \parencite{wasserstein2016asa}.

\subsection{Diagnostic-Pattern Audit Summary}
\label{subsec:failure_taxonomy_main}
The categorical pattern labels are used as aggregate descriptive diagnostics rather than item-level causal ground truth. The operational rules and validation audit are reported in Appendix~\ref{app:failure_taxonomy_rules} and Appendix~\ref{app:human_failure_validation}; the main text relies primarily on answer/evidence metrics and uses label counts only as supporting evidence.
\section{Experiments}

The experiments are organized around the main diagnostic chain. The body reports the primary four-condition matrix, ONCU validity behavior, realistic multi-hop robustness, model-family extension, matched dense/hybrid sensitivity, and controlled length--position scaling. The appendix retains the larger set of confidence intervals, diagnostic-pattern audit details, raw-versus-clipped rows, retrieval sweeps, and external validation tables.

\subsection{Experimental Setup}
The setup applies the four-condition protocol to controlled and realistic QA settings. The claims are scoped to local open-weight models, reconstructed oracle-compatible QA protocols, and the retriever families tested here.

The primary ONCU-compatible model-generation evidence consists of 18,000 fixed-condition predictions across five local open-weight models from the Qwen, Gemma, Llama, and Mistral families. The balanced core matrix evaluates Qwen2.5-14B \parencite{qwen2025qwen25}, Qwen3-14B \parencite{qwen2025qwen3}, and Gemma3-12B \parencite{gemma2025gemma3} on Controlled-ONCU-safe16K-200 and HotpotQA-ONCU-200 under the four diagnostic conditions, yielding 4,800 predictions. The same three-model panel is then evaluated on 2WikiMultiHopQA-ONCU-500, yielding 6,000 additional predictions. A model-family extension repeats the same protocol for Llama3.1-8B \parencite{meta2024llama31} and Mistral-Small3.1-24B \parencite{mistral2025small31} on all three ONCU-compatible components, yielding 7,200 predictions. The 18,000 predictions are therefore the primary ONCU-compatible evidence base; all other prediction counts reported below are auxiliary sensitivity or audit runs.

The controlled safe16K subset contains samples drawn from 4K, 8K, and 16K contexts. The 32K setting is excluded from the core model comparison because a 32K context plus task instructions, question text, and structured-output constraints can approach backend context limits and introduce truncation-related confounds. This restriction reduces truncation confounds when comparing similarly sized local models.

The ONCU-compatible model-generation experiments use deterministic decoding with temperature set to 0, maximum generation length set to 1024 tokens, and an explicit Ollama context-window configuration of 32,768 tokens \parencite{ollama2026}. The main retrieved-evidence condition uses lexical retrieval with top-$k=3$, chunk size 220, and overlap 40. For Qwen3-14B, thinking output is disabled at the API level to preserve the same structured-output contract used for the other models; the task instructions and diagnostic conditions remain unchanged.

Structured-output parse failures are retained and scored as incorrect. Four of the six final 200-sample primary runs have zero parse failures. Qwen3-14B on HotpotQA-ONCU-200 has one parse failure out of 800 predictions. Gemma3-12B on Controlled-ONCU-safe16K-200 has six parse failures out of 800 predictions, all under the full-context condition. Gemma3-12B on HotpotQA-ONCU-200 has zero parse failures. In the primary 2WikiMultiHopQA-ONCU-500 runs, Qwen2.5-14B and Gemma3-12B each have one parse failure out of 2000 predictions, while Qwen3-14B has zero parse failures. In the Llama/Mistral extension, the controlled runs have zero parse failures; the HotpotQA runs have two parse failures for Llama3.1-8B and one for Mistral-Small3.1-24B; and the 2WikiMultiHopQA runs have three parse failures for Llama3.1-8B and one for Mistral-Small3.1-24B. Keeping parse failures in the denominator avoids selectively discarding difficult outputs and keeps the model comparisons conservative.

Auxiliary runs are used only as sensitivity or supporting evidence. They include BABILong-200 and RULER-lite-240 external answer-performance checks, retrieval-only family and budget audits, complete dense@16 and hybrid@16 four-condition ONCU reruns, broader reader-facing retriever-family sweeps, a five-model cross-encoder reranking audit, and a controlled length--position scaling extension. These runs test whether the condition-level diagnosis is stable under stronger retrieval, additional model families, external long-context tasks, and known evidence-position structure; they do not redefine the primary ONCU-compatible evidence base.

\subsection{Sensitivity Analyses as Threats Addressed}
\label{subsec:threats_addressed}
The auxiliary analyses are treated as sensitivity and supporting evidence for the main condition-level diagnosis, not as separate benchmark contributions. Table~\ref{tab:threats_addressed_by_sensitivity} summarizes the threat each audit addresses, the corresponding check, and whether it changes the main interpretation.

\begin{table*}[!tbp]
\renewcommand{\arraystretch}{1.08}
\caption{Threats Addressed by Sensitivity and Supporting Analyses. These audits qualify the interpretation of the four-condition protocol without changing the central claim: evidence-use patterns differ across controlled and realistic settings under the tested evidence conditions.}
\label{tab:threats_addressed_by_sensitivity}
\centering
\scriptsize
\setlength{\tabcolsep}{2.5pt}
\resizebox{\textwidth}{!}{%
\begin{tabular}{p{0.24\textwidth}p{0.25\textwidth}p{0.34\textwidth}p{0.13\textwidth}}
\toprule
Threat to interpretation & Audit / sensitivity check & Conclusion under tested conditions & Changes main claim? \\
\midrule
Lexical@3 retrieval is too weak to represent retrieval-conditioned input.
& Retrieval-budget audit, retriever-family audit, dense@16/hybrid@16 matched ONCU sensitivity, and cross-encoder reranking appendix.
& Stronger retrieval narrows some gaps and improves retrieved performance, but does not overturn the full-context-over-retrieved direction in the tested realistic multi-hop protocol.
& No; it qualifies retrieval scope. \\

The result is specific to one model family.
& Five-model panel: Qwen2.5-14B, Qwen3-14B, Gemma3-12B, Llama3.1-8B, and Mistral-Small3.1-24B.
& The controlled-versus-realistic diagnostic pattern remains visible across the tested local open-weight families.
& No. \\

HotpotQA denominator-valid groups are too narrow for unconditional ONCU claims.
& Denominator-valid audit, HotpotQA-500 robustness, and denominator-free answer/evidence metrics.
& ONCU is interpreted only on oracle-improving groups; sample-level answer/evidence metrics support the same direction without denominator filtering.
& No; it limits the ONCU scope. \\

External long-context tasks might imply a different interpretation.
& BABILong-200 and RULER-lite-240 answer-performance checks.
& These checks are supportive external validation only, because they do not instantiate the full oracle-reference protocol.
& No; no ONCU claim is made from them. \\

Failure labels may be artifacts of the automatic taxonomy.
& Human validation of failure-type assignment.
& The taxonomy is used as aggregate descriptive support, while the main claims rely on answer/evidence metrics and denominator-valid ONCU.
& No. \\
\bottomrule
\end{tabular}
}
\end{table*}
\FloatBarrier

\subsection{Retrieval Baseline Interpretation and Sensitivity Preview}
\label{subsec:retrieval_baseline_preview}
The main retrieved-evidence rows use lexical@3 as a matched diagnostic condition. Retrieval strength is then audited in three ways: complete dense@16/hybrid@16 four-condition ONCU reruns in the main text, broader reader-facing sweeps in Appendix~\ref{app:reader_facing_validation}, and five-model cross-encoder reranking in Appendix~\ref{app:ce_reranking_audit}. These analyses show how the diagnosis changes when the retrieved input is strengthened, and they are summarized in Table~\ref{tab:threats_addressed_by_sensitivity}.

\subsection{Main 200-Sample Answer and Evidence Results}
The core 200-sample matrix shows a task-dependent diagnostic pattern. Controlled examples favor compact retrieved or oracle evidence over full context, while HotpotQA-derived examples favor full context over the fixed retrieved condition. These first results use denominator-free answer and evidence scores over all evaluated samples; ONCU is interpreted only after the denominator-valid group audit in the next subsection.

Table~\ref{tab:sci200_main_results} reports the final 200-sample results. The controlled setting reveals a consistent full-context recovery gap across all three models. Qwen2.5-14B, Qwen3-14B, and Gemma3-12B obtain full-context relaxed F1 values of 0.538, 0.526, and 0.514, respectively. In contrast, their retrieved-evidence relaxed F1 values are 0.975, 0.993, and 0.845, and their oracle-evidence reference scores are 0.903, 0.995, and 0.990. In the controlled safe16K setting, all three models can therefore answer from compact or isolated evidence substantially better than from the same evidence embedded in a long controlled context.

The HotpotQA-derived setting shows the opposite pattern under the evaluated retrieved input. For all three models, full-context answering outperforms retrieved-evidence evaluation. Qwen2.5-14B obtains full-context relaxed F1 of 0.733 compared with 0.590 under retrieved evidence; Qwen3-14B obtains 0.689 compared with 0.558; and Gemma3-12B obtains 0.668 compared with 0.561. Evidence F1 values show the same direction: full-context evidence F1 is consistently higher than retrieved-evidence F1. The fixed compact retrieved condition can therefore remove or under-rank supporting facts required for the tested HotpotQA multi-hop setting. The diagnostic conclusion is about evidence-chain coverage under the evaluated retriever families and budgets. The reader-facing retrieval validation in Appendix~\ref{app:reader_facing_validation} shows that stronger or larger-budget retrieved inputs can improve retrieved answer F1, but the best retrieved configuration is not uniform across models and datasets and must be interpreted together with distractor exposure.

\begin{table*}[!tbp]
\renewcommand{\arraystretch}{1.12}
\caption{Final 200-Sample Core Diagnostic Results. Each row summarizes 200 examples evaluated under four core conditions. Full denotes full-context input, Ret. denotes retrieved evidence, and Oracle-ref. denotes the oracle-evidence reference. Parse errors are retained and scored as incorrect.}
\label{tab:sci200_main_results}
\centering
\scriptsize
\setlength{\tabcolsep}{3pt}
\begin{tabular}{llccccccc}
\toprule
Model & Dataset & No-Ev. F1 & Full F1 & Ret. F1 & Oracle-ref. F1 & Full Ev.F1 & Ret. Ev.F1 & Parse Err. \\
\midrule
Qwen2.5-14B & Controlled-safe16K-200 & 0.000 & 0.538 & 0.975 & 0.903 & 0.596 & 0.975 & 0/800 \\
Qwen2.5-14B & HotpotQA-ONCU-200 & 0.241 & 0.733 & 0.590 & 0.767 & 0.688 & 0.478 & 0/800 \\
Qwen3-14B & Controlled-safe16K-200 & 0.008 & 0.526 & 0.993 & 0.995 & 0.706 & 0.972 & 0/800 \\
Qwen3-14B & HotpotQA-ONCU-200 & 0.301 & 0.689 & 0.558 & 0.795 & 0.680 & 0.469 & 1/800 \\
Gemma3-12B & Controlled-safe16K-200 & 0.000 & 0.514 & 0.845 & 0.990 & 0.673 & 0.793 & 6/800 \\
Gemma3-12B & HotpotQA-ONCU-200 & 0.291 & 0.668 & 0.561 & 0.791 & 0.576 & 0.406 & 0/800 \\
\bottomrule
\end{tabular}
\end{table*}

\subsection{Denominator-Valid Group Audit and Sensitivity Checks}
\label{subsec:denominator_valid_group_audit}

ONCU validity is treated as a result in its own right rather than as a hidden preprocessing step. The valid/invalid group audit in Table~\ref{tab:denominator_validity_audit} and the HotpotQA filter-sensitivity table below specify where the denominator supports a recovered-advantage interpretation. HotpotQA ONCU is therefore conditional, not an unconditional HotpotQA score. It applies only to oracle-improving metadata groups, while the dataset-level HotpotQA direction is supported by denominator-free answer and evidence metrics over all evaluated samples.

Table~\ref{tab:denominator_validity_audit} gives the valid-group counts for the main 200-sample matrix and the larger HotpotQA-500 robustness runs before the ONCU table is interpreted. The controlled safe16K benchmark has broad denominator coverage: all three models have at least 133 valid groups out of 134. HotpotQA-ONCU-200 is more restrictive, with 28--29 valid groups out of 38.

The larger HotpotQA-500 runs audit whether the valid-group coverage changes with sample size under the same metadata grouping scheme and evaluation contract. Unique HotpotQA groups increase from 38 to 46. Valid groups increase from 29 to 43 for Qwen2.5-14B, from 28 to 41 for Qwen3-14B, and from 28 to 38 for Gemma3-12B. The remaining invalid groups have the same denominator status: the oracle-evidence reference does not exceed the no-evidence baseline under relaxed answer F1. This means the denominator-invalid groups are not parse failures or failed evidence mappings; they are groups for which the model is already comparatively answerable without evidence, or the isolated oracle snippet does not create additional recoverable advantage.

\begin{table*}[!tbp]
\renewcommand{\arraystretch}{1.10}
\caption{Denominator-Validity Audit for ONCU Aggregation. Valid groups satisfy $S_{\mathrm{oracle}}>S_{\mathrm{no}}$ under the relaxed-F1 score field. The HotpotQA-500 runs are larger-sample robustness checks, not replacements for the balanced 200-sample matrix.}
\label{tab:denominator_validity_audit}
\centering
\scriptsize
\setlength{\tabcolsep}{3pt}
\begin{tabular}{llrrrrp{0.31\textwidth}}
\toprule
Model & Dataset & Samples & Total Groups & Valid Groups & Invalid Groups & Interpretation \\
\midrule
Qwen2.5-14B & Controlled-safe16K & 200 & 134 & 134 & 0 & Broad denominator coverage. \\
Qwen3-14B & Controlled-safe16K & 200 & 134 & 133 & 1 & Broad denominator coverage. \\
Gemma3-12B & Controlled-safe16K & 200 & 134 & 133 & 1 & Broad denominator coverage. \\
Qwen2.5-14B & HotpotQA-ONCU & 200 & 38 & 29 & 9 & ONCU applies to oracle-improving groups. \\
Qwen3-14B & HotpotQA-ONCU & 200 & 38 & 28 & 10 & ONCU applies to oracle-improving groups. \\
Gemma3-12B & HotpotQA-ONCU & 200 & 38 & 28 & 10 & ONCU applies to oracle-improving groups. \\
Qwen2.5-14B & HotpotQA-ONCU & 500 & 46 & 43 & 3 & Larger-sample audit preserves the direction. \\
Qwen3-14B & HotpotQA-ONCU & 500 & 46 & 41 & 5 & Larger-sample audit preserves the direction. \\
Gemma3-12B & HotpotQA-ONCU & 500 & 46 & 38 & 8 & Larger-sample audit preserves the direction. \\
\bottomrule
\end{tabular}
\end{table*}

Table~\ref{tab:hotpotqa_denominator_filter_sensitivity} makes the HotpotQA-ONCU-200 denominator boundary explicit at main-text level. The invalid groups are all comparison-type metadata groups under the current grouping scheme. Their answer-type mix is dominated by entity questions, with smaller numbers of yes/no, string, and numeric answers. The raw invalid-group relaxed-F1 columns show the scores before ONCU filtering; they make the denominator boundary visible in the main results. The final three columns compare three ways to summarize the full-minus-retrieved direction: a sample-level relaxed-F1 contrast over all examples, the usual unweighted group-averaged ONCU contrast over denominator-valid groups, and a denominator-weighted ONCU contrast over the same valid groups.

\begin{table*}[!tbp]
\renewcommand{\arraystretch}{1.10}
\caption{HotpotQA-ONCU-200 Denominator-Filter Sensitivity. Invalid groups are those with $S_{\mathrm{oracle}}\leq S_{\mathrm{no}}$. Raw invalid-group F1 is reported as No/Full/Ret./Oracle. The answer-type profile counts samples inside invalid groups; E=entity, N=number, S=string, Y=yes/no. The three contrast columns report Full/Ret./$\Delta$ for sample-level relaxed F1, unweighted valid-group ONCU, and denominator-weighted valid-group ONCU.}
\label{tab:hotpotqa_denominator_filter_sensitivity}
\centering
\scriptsize
\setlength{\tabcolsep}{2.0pt}
\resizebox{\textwidth}{!}{%
\begin{tabular}{lccp{0.16\textwidth}p{0.17\textwidth}cp{0.12\textwidth}p{0.12\textwidth}p{0.12\textwidth}}
\toprule
Model & Valid & Invalid & Invalid group type & Invalid answer profile & Raw invalid F1 & Sample F1 & Unweighted ONCU & Denom.-weighted ONCU \\
\midrule
Qwen2.5-14B & 29/38 & 9 & 9 comparison groups; all oracle$\leq$no-evidence & E:8, N:1, S:1, Y:3 & 0.698/0.753/0.685/0.698 & 0.733/0.590/+0.143 & 0.906/0.639/+0.267 & 0.948/0.656/+0.292 \\
Qwen3-14B & 28/38 & 10 & 10 comparison groups; all oracle$\leq$no-evidence & E:12, N:1, S:1, Y:2 & 0.742/0.729/0.825/0.703 & 0.689/0.558/+0.131 & 0.787/0.557/+0.231 & 0.820/0.565/+0.254 \\
Gemma3-12B & 28/38 & 10 & 10 comparison groups; all oracle$\leq$no-evidence & E:9, N:1, S:2, Y:4 & 0.775/0.651/0.754/0.775 & 0.668/0.561/+0.107 & 0.719/0.536/+0.183 & 0.764/0.585/+0.179 \\
\bottomrule
\end{tabular}%
}
\end{table*}

This audit sharpens the HotpotQA interpretation. First, the denominator-invalid groups are concentrated in a recognizable metadata region instead of arising from parse failures or unavailable oracle passages. Second, sample-level answer F1 over all HotpotQA-ONCU-200 examples supports full context over retrieved evidence for all three models, independently of ONCU filtering. Third, both unweighted valid-group ONCU and denominator-weighted valid-group ONCU preserve the same full-over-retrieved direction. The resulting claim is intentionally two-layered: sample-level answer and evidence metrics support full-over-retrieved behavior on the evaluated HotpotQA-ONCU samples, and ONCU supports the same direction only on oracle-improving groups.

\subsection{Oracle-Reference Normalized Context Utilization}
After denominator validity is made explicit, ONCU supports the same task-dependent diagnostic pattern. The group-normalized relaxed-F1 ONCU table is interpreted in light of Tables~\ref{tab:denominator_validity_audit} and~\ref{tab:hotpotqa_denominator_filter_sensitivity}. The values are diagnostics over oracle-improving groups, not unconditional dataset-level scores.

The score columns in Table~\ref{tab:sci200_oncu_results} are group averages, so they may differ slightly from the example-level means in Table~\ref{tab:sci200_main_results}. ONCU is averaged only over metadata groups for which the oracle-evidence reference condition exceeds the no-evidence condition.

The controlled results show broad denominator coverage and a stable full-context recovery gap. Full-context ONCU is 0.583 for Qwen2.5-14B, 0.535 for Qwen3-14B, and 0.515 for Gemma3-12B. In contrast, retrieved-evidence ONCU is 0.981, 0.994, and 0.842, respectively. In this setting, both ONCU and denominator-free example-level F1 point in the same direction: compact evidence recovers much more of the oracle-reference advantage than the full long context.

The HotpotQA-derived results show the reverse relationship on oracle-improving groups. Full-context ONCU is 0.906 for Qwen2.5-14B, 0.787 for Qwen3-14B, and 0.719 for Gemma3-12B, while retrieved-evidence ONCU is 0.639, 0.557, and 0.536. This conditional ONCU result is aligned with the denominator-free evidence already reported in Table~\ref{tab:sci200_main_results}: raw answer F1 and evidence F1 both favor full context over retrieved evidence for all three models. The broader HotpotQA conclusion therefore rests on two compatible layers: sample-level answer/evidence metrics over all evaluated examples, and ONCU over denominator-valid oracle-improving groups.

\begin{table*}[!tbp]
\renewcommand{\arraystretch}{1.12}
\caption{Final 200-Sample ONCU-Relaxed-F1 Results. $S_{\mathrm{full}}$ and $S_{\mathrm{ret}}$ denote group-averaged relaxed F1 for the full-context and retrieved-evidence conditions.}
\label{tab:sci200_oncu_results}
\centering
\scriptsize
\setlength{\tabcolsep}{3pt}
\begin{tabular}{llccccccc}
\toprule
Model & Dataset & Valid Groups & $S_{\mathrm{no}}$ & $S_{\mathrm{oracle}}$ & $S_{\mathrm{full}}$ & Full ONCU & $S_{\mathrm{ret}}$ & Ret. ONCU \\
\midrule
Qwen2.5-14B & Controlled-safe16K-200 & 134 & 0.000 & 0.909 & 0.546 & 0.583 & 0.981 & 0.981 \\
Qwen2.5-14B & HotpotQA-ONCU-200 & 29 & 0.212 & 0.793 & 0.763 & 0.906 & 0.593 & 0.639 \\
Qwen3-14B & Controlled-safe16K-200 & 133 & 0.008 & 1.000 & 0.535 & 0.535 & 0.994 & 0.994 \\
Qwen3-14B & HotpotQA-ONCU-200 & 28 & 0.236 & 0.804 & 0.702 & 0.787 & 0.557 & 0.557 \\
Gemma3-12B & Controlled-safe16K-200 & 133 & 0.000 & 0.996 & 0.515 & 0.515 & 0.842 & 0.842 \\
Gemma3-12B & HotpotQA-ONCU-200 & 28 & 0.215 & 0.787 & 0.652 & 0.719 & 0.550 & 0.536 \\
\bottomrule
\end{tabular}
\end{table*}

\subsection{Aggregate Raw-vs-Clipped ONCU Audit}
\label{subsec:raw_clipped_audit}

The main controlled-versus-realistic split is not an artifact of clipping. The raw-ratio audit in Appendix~\ref{app:raw_clipped_audit} shows one aggregate above-oracle case, Qwen2.5-14B on Controlled-safe16K under retrieved evidence, with a raw ratio of 1.079 before clipping. Clipped values support the recovered-fraction summary, while raw values remain the audit trail for above-oracle or below-baseline behavior.

\subsection{Statistical and Failure-Audit Summary}
\label{subsec:statistical_failure_summary_main}
Bootstrap intervals, paired contrasts, and aggregate diagnostic-pattern summaries support the mean-table conclusions. Detailed numeric audit tables are placed in Appendix~\ref{app:bootstrap_ci}, Appendix~\ref{app:failure_breakdown}, and Appendix~\ref{app:statistical_checks} so that the main text does not treat each audit as an independent result.
\subsection{Cross-Model Findings}
Stronger isolated-evidence performance does not automatically imply stronger full-context utilization. In the 200-sample controlled setting, Qwen3-14B and Gemma3-12B reach oracle-reference relaxed F1 near 1.0, but their full-context ONCU values remain 0.535 and 0.515. Qwen2.5-14B has lower oracle-reference F1 but slightly higher full-context ONCU at 0.583. This within-protocol comparison separates evidence-use behavior from isolated-evidence answer accuracy.

The Gemma3-12B controlled result adds a cross-family perspective. Gemma3-12B has a high oracle-reference relaxed F1 of 0.990, but its retrieved-evidence ONCU is 0.842, lower than the Qwen-family values above 0.98. ONCU separates full-context utilization behavior from model-specific sensitivity to compact retrieved evidence, a distinction that aggregate answer accuracy would obscure.

\subsection{Dataset-Dependent Diagnostic Patterns}
Evidence-use behavior is dataset- and pipeline-dependent. In Controlled-ONCU, retrieved-evidence evaluation dominates full-context answering in both example-level metrics and broad-coverage ONCU. In HotpotQA-ONCU, denominator-free answer/evidence metrics favor full context, and ONCU agrees over oracle-improving groups. The HotpotQA retrieved-evidence result points to evidence-chain coverage under the evaluated retrieval conditions.

This difference illustrates why ONCU should be interpreted together with evidence F1 and dataset structure. A low full-context ONCU may reflect reduced use of available evidence, whereas a low retrieved-evidence ONCU may reflect pre-reader evidence coverage rather than answer generation alone. The HotpotQA results also show why no-evidence normalization is necessary: all three models obtain non-trivial no-evidence relaxed F1, indicating that parametric knowledge or question priors can otherwise be mistaken for context utilization.

\subsection{Three-Model Controlled Length--Position Scaling}
\label{subsec:controlled_scaling}

The controlled full-context deficit varies systematically with context length and evidence position. The scaling extension uses the same 3,200 generated samples for Qwen2.5-14B, Qwen3-14B, and Gemma3-12B, crossing four context lengths, ten decile evidence-position buckets, four reasoning types, four distractor-similarity settings, and five seeds per cell. This extension serves as controlled structural evidence for the controlled benchmark, not as a replacement for the balanced cross-dataset matrix.

Table~\ref{tab:controlled_scaling_three_model_means} summarizes the length effect. Across all three models, mean full-context ONCU declines sharply as context length increases. Qwen2.5-14B falls from 0.999 at 4K to 0.163 at 32K. Qwen3-14B falls from 0.980 to 0.150. Gemma3-12B starts lower at 4K, with mean full-context ONCU of 0.791, but converges to the same low 32K regime, with mean full-context ONCU of 0.157. The controlled full-context deficit is therefore not limited to one model family.

\begin{table*}[!tbp]
\renewcommand{\arraystretch}{1.10}
\caption{Three-Model Controlled Length Scaling. Each full-context cell reports clipped ONCU-Relaxed-F1 averaged over the ten evidence-position buckets for a fixed context length. Retrieved-evidence ONCU is included as an internal compact-evidence competence check.}
\label{tab:controlled_scaling_three_model_means}
\centering
\scriptsize
\setlength{\tabcolsep}{4pt}
\begin{tabular}{lcccccccc}
\toprule
\multirow{2}{*}{Model}
& \multicolumn{4}{c}{Full-context ONCU}
& \multicolumn{4}{c}{Retrieved-evidence ONCU} \\
\cmidrule(lr){2-5}\cmidrule(lr){6-9}
& 4K & 8K & 16K & 32K & 4K & 8K & 16K & 32K \\
\midrule
Qwen2.5-14B & 0.999 & 0.599 & 0.301 & 0.163 & 1.000 & 1.000 & 1.000 & 1.000 \\
Qwen3-14B   & 0.980 & 0.618 & 0.300 & 0.150 & 0.999 & 0.999 & 1.000 & 1.000 \\
Gemma3-12B  & 0.791 & 0.543 & 0.307 & 0.157 & 0.847 & 0.843 & 0.836 & 0.849 \\
\bottomrule
\end{tabular}
\end{table*}

Table~\ref{tab:controlled_scaling_three_model_32k_position} reports the 32K position profile, where the length effect is most severe. The qualitative pattern is shared across models: early and middle positions recover little oracle-referenced advantage, while the final evidence decile remains the strongest. Qwen2.5-14B and Qwen3-14B show an especially sharp recency-skewed collapse: at 32K, clipped full-context ONCU is near zero through most early and middle deciles, partially recovers at \texttt{pos\_08}, and reaches 1.000 at \texttt{pos\_09}. Gemma3-12B shows a smoother but directionally consistent profile, with low early and middle ONCU values, partial recovery at \texttt{pos\_08}, and its strongest 32K value at \texttt{pos\_09}.

\begin{table*}[!tbp]
\renewcommand{\arraystretch}{1.06}
\caption{32K Full-Context ONCU by Evidence Position. Each cell reports clipped ONCU-Relaxed-F1 at the longest controlled context length. The full length--position grid is released in the controlled scaling artifacts.}
\label{tab:controlled_scaling_three_model_32k_position}
\centering
\scriptsize
\setlength{\tabcolsep}{3pt}
\begin{tabular}{lcccccccccc}
\toprule
Model
& \texttt{pos\_00} & \texttt{pos\_01} & \texttt{pos\_02} & \texttt{pos\_03} & \texttt{pos\_04}
& \texttt{pos\_05} & \texttt{pos\_06} & \texttt{pos\_07} & \texttt{pos\_08} & \texttt{pos\_09} \\
\midrule
Qwen2.5-14B & 0.013 & 0.000 & 0.007 & 0.047 & 0.048 & 0.055 & 0.020 & 0.007 & 0.431 & 1.000 \\
Qwen3-14B   & 0.000 & 0.000 & 0.000 & 0.000 & 0.000 & 0.000 & 0.000 & 0.000 & 0.504 & 1.000 \\
Gemma3-12B  & 0.007 & 0.030 & 0.068 & 0.072 & 0.000 & 0.107 & 0.028 & 0.055 & 0.454 & 0.750 \\
\bottomrule
\end{tabular}
\end{table*}

The retrieved-evidence condition provides the key internal validity check. For Qwen2.5-14B and Qwen3-14B, retrieved-evidence ONCU remains essentially saturated across context lengths. Gemma3-12B also benefits substantially from compact evidence, but its retrieved-evidence ONCU remains around 0.84 rather than 1.00. This difference is diagnostically important. It shows that all three models show a full-context length--position utilization collapse, while Gemma3-12B also retains an additional compact-evidence limitation after localization has been simplified.

Failure-type analysis is consistent with this interpretation. In the Gemma3-12B scaling run, the retrieved-evidence condition has only 21 evidence-localization failures out of 3,200 examples, but still has 802 evidence-integration failures and 823 answer-conversion failures. The oracle-evidence condition has 1,747 successes and 1,453 answer-conversion failures. Gemma3-12B's lower retrieved-evidence ONCU is therefore not primarily caused by missing compact evidence; it reflects a reader-side conversion and integration limitation. In contrast, the Qwen-family models recover nearly all oracle-reference advantage when the evidence is compactly supplied, so their controlled scaling deficit is more directly attributable to full-context evidence localization and utilization.

We treat the scaling extension as a three-model mechanism audit. It is consistent with prior observations that long-context models can be sensitive to the position of relevant information, but the ONCU framing asks a sharper question: whether a model recovers the same oracle-evidence advantage when evidence is embedded in the full context versus compactly supplied. The main conclusion is not that all models behave identically. Rather, the shared qualitative trend is that full-context ONCU degrades with length and evidence position, while the recovered compact-evidence advantage differs by model family.

\subsection{Supporting Retrieval and External-Validation Evidence}
\label{subsec:supporting_evidence_summary}
The retrieval-only, reader-facing, cross-encoder, BABILong, and RULER-lite analyses are interpreted as supporting evidence summarized in Table~\ref{tab:threats_addressed_by_sensitivity}. They are not used to redefine the central claim. Their shared role is to test whether the task-dependent diagnostic pattern survives stronger retrieval conditions, additional model families, external answer-performance checks, and failure-label validation. Detailed retrieval-budget and retrieval-only retriever-family audits are reported in Appendices~\ref{app:retrieval_budget} and~\ref{app:retrieval_ablation}. BABILong-200 and RULER-lite-240 external-validation tables are reported in Appendices~\ref{app:babilong_validation} and~\ref{app:ruler_lite_validation}. Full audit tables are preserved in Appendix~\ref{app:supplementary_audits}.

\subsection{HotpotQA-500 Robustness and Valid-Group Audit}
The HotpotQA-500 robustness run tests whether the 200-sample realistic multi-hop pattern is a small-sample artifact. It uses the same top-$k=3$ diagnostic protocol for Qwen2.5-14B, Qwen3-14B, and Gemma3-12B; it changes sample size, not the task definition or retrieval condition.

\begin{table*}[!tbp]
\renewcommand{\arraystretch}{1.12}
\caption{HotpotQA-500 Robustness and Valid-Group Audit for Qwen2.5-14B, Qwen3-14B, and Gemma3-12B. The 500-sample runs preserve the sample-level full-context-over-retrieved direction and increase valid ONCU groups. Bracketed values are 95\% bootstrap confidence intervals for ONCU-Relaxed-F1 over denominator-valid groups.}
\label{tab:hotpotqa500_robustness_all}
\centering
\scriptsize
\setlength{\tabcolsep}{2.2pt}
\begin{tabular}{llccccp{0.29\textwidth}}
\toprule
Model & Setting & Samples & Valid Groups & Full / Ret. F1 & Full / Ret. Ev.F1 & Full / Ret. ONCU [95\% CI] \\
\midrule
Qwen2.5-14B & HotpotQA-200 & 200 & 29 & 0.733 / 0.590 & 0.688 / 0.478 & 0.906 [0.843, 0.958] / 0.639 [0.511, 0.755] \\
Qwen2.5-14B & HotpotQA-500 & 500 & 43 & 0.733 / 0.600 & 0.676 / 0.465 & 0.845 [0.769, 0.909] / 0.657 [0.575, 0.733] \\
Qwen3-14B & HotpotQA-200 & 200 & 28 & 0.689 / 0.558 & 0.680 / 0.469 & 0.787 [0.679, 0.881] / 0.557 [0.441, 0.675] \\
Qwen3-14B & HotpotQA-500 & 500 & 41 & 0.715 / 0.582 & 0.699 / 0.456 & 0.787 [0.710, 0.854] / 0.593 [0.510, 0.675] \\
Gemma3-12B & HotpotQA-200 & 200 & 28 & 0.668 / 0.561 & 0.576 / 0.406 & 0.719 [0.604, 0.819] / 0.536 [0.428, 0.645] \\
Gemma3-12B & HotpotQA-500 & 500 & 38 & 0.675 / 0.576 & 0.587 / 0.399 & 0.671 [0.578, 0.759] / 0.483 [0.385, 0.579] \\
\bottomrule
\end{tabular}
\end{table*}

Table~\ref{tab:hotpotqa500_robustness_all} preserves the full comparison. The larger run keeps the same qualitative pattern: full-context relaxed F1, evidence F1, and ONCU remain higher than retrieved-evidence values for all three models. The valid-group audit also improves rather than weakens the denominator story: the number of unique metadata groups increases from 38 to 46, and the valid ONCU group count rises for every model. This result supports the same two-layer interpretation used throughout the paper: denominator-free sample metrics support the dataset-level direction, and ONCU supports the same direction on oracle-improving groups.

\subsection{2WikiMultiHopQA-ONCU-500 Realistic Multi-hop Validation}
\label{subsec:twowiki500_results}

\input{twowiki_results_paragraph}

\input{twowiki_main_results_table}

\input{twowiki_oncu_results_table}

\input{twowiki_failure_breakdown_table}

\subsection{Model-Family Extension: Llama and Mistral}
\label{subsec:model_family_extension}
The model-family extension tests whether the task-dependent diagnostic pattern is specific to the original Qwen/Gemma panel. Llama3.1-8B and Mistral-Small3.1-24B are evaluated with the same processed inputs, evidence conditions, decoding policy, answer contract, lexical retrieval setting, and scoring pipeline.

\begin{table*}[!tbp]
\renewcommand{\arraystretch}{1.10}
\caption{Model-Family Extension Results. The extension adds Llama3.1-8B and Mistral-Small3.1-24B to the same four-condition ONCU-compatible protocol. F1 columns are example-level relaxed answer F1. ONCU columns are group-averaged clipped ONCU computed from relaxed answer F1 over denominator-valid metadata groups.}
\label{tab:model_family_extension_results}
\centering
\scriptsize
\setlength{\tabcolsep}{2.4pt}
\begin{tabular}{llrrrrrrrrr}
\toprule
Model & Dataset & Valid & No-Ev. & Full & Ret. & Oracle-ref. & Full & Ret. & Full & Ret. \\
 & & Groups & F1 & F1 & F1 & F1 & ONCU & ONCU & Ev.F1 & Ev.F1 \\
\midrule
Llama3.1-8B & Controlled-safe16K-200 & 134 & 0.000 & 0.729 & 0.930 & 1.000 & 0.760 & 0.938 & 0.684 & 0.750 \\
Mistral-Small3.1-24B & Controlled-safe16K-200 & 134 & 0.022 & 0.621 & 0.995 & 1.000 & 0.625 & 0.996 & 0.828 & 0.984 \\
Llama3.1-8B & HotpotQA-ONCU-200 & 30 & 0.282 & 0.649 & 0.552 & 0.761 & 0.742 & 0.568 & 0.497 & 0.435 \\
Mistral-Small3.1-24B & HotpotQA-ONCU-200 & 30 & 0.352 & 0.744 & 0.592 & 0.830 & 0.809 & 0.539 & 0.717 & 0.511 \\
Llama3.1-8B & 2WikiMultiHopQA-ONCU-500 & 33 & 0.282 & 0.472 & 0.361 & 0.725 & 0.371 & 0.254 & 0.470 & 0.333 \\
Mistral-Small3.1-24B & 2WikiMultiHopQA-ONCU-500 & 35 & 0.278 & 0.625 & 0.409 & 0.799 & 0.596 & 0.308 & 0.727 & 0.358 \\
\bottomrule
\end{tabular}
\end{table*}

Table~\ref{tab:model_family_extension_results} preserves the condition-level details. Both added models keep the controlled synthetic pattern: compact retrieved evidence recovers more oracle-reference advantage than full long context. In the realistic multi-hop settings, the direction reverses: full context remains above deterministic retrieved input. The extension therefore strengthens model-family robustness within the tested local open-weight scope without claiming coverage of proprietary frontier systems or larger learned-retrieval pipelines.

\subsection{Matched Retriever-Family ONCU Sensitivity}
\label{subsec:retriever_family_oncu_sensitivity}
The dense@16 and hybrid@16 sensitivity experiment is the main-text retrieval-strength audit because it reruns the complete four-condition protocol on the same model--dataset pairs. It therefore preserves the no-evidence, full-context, retrieved-evidence, and oracle-evidence references needed for ONCU.

\input{retriever_family_oncu_sensitivity_table}

Table~\ref{tab:retriever_family_oncu_sensitivity} shows that stronger retrieved inputs improve several retrieved-evidence rows, but do not overturn the realistic multi-hop diagnosis under the tested settings. On HotpotQA-ONCU-200, full-context F1 and ONCU remain higher than dense@16 and hybrid@16 retrieved values for all three primary models. On 2WikiMultiHopQA-ONCU-500, the retrieved condition improves substantially relative to lexical@3, yet full-context ONCU remains higher for all three models, with Gemma3-12B nearly closing the gap under dense@16. The conclusion is therefore not that retrieval is intrinsically worse; it is that retrieval strength changes the magnitude of the observed gap while the matched protocol still separates pre-reader evidence availability from reader-side conversion.

\section{Discussion}
The main scientific claim is that final accuracy, retrieval coverage, and citation overlap do not directly identify the behavioral evidence-utilization contrast studied here. That contrast becomes observable only through matched evidence-availability conditions whose denominator-valid scope is reported before condition-level ONCU values are interpreted. The no-evidence condition estimates answerability that should not be credited to the supplied context; the oracle-evidence reference estimates the recoverable advantage of isolated evidence; and the full-context and retrieved-evidence conditions test whether that advantage survives the actual input regime. ONCU is the conditional estimator used for this contrast, while the accompanying audits determine whether the denominator, score field, and companion evidence support the interpretation.

The tested patterns should therefore be interpreted at the level of evidence presentation, retrieval, and reading, rather than as a simple long-context-versus-RAG comparison. In controlled synthetic settings, compact or isolated evidence is often sufficient for correct answering, but the same evidence embedded in a long context yields substantially lower recovered oracle-reference advantage. In the tested realistic multi-hop reconstructions, full context often outperforms the evaluated retrieved inputs, indicating that pre-reader evidence-chain coverage can determine the downstream contrast. The central result is not that full context or retrieval is uniformly better, but that evidence-use behavior is task- and pipeline-dependent under the evaluated conditions.

The auxiliary analyses support this interpretation without replacing the core protocol. Table~\ref{tab:threats_addressed_by_sensitivity} organizes them as threats addressed: model-family coverage, retrieval strength, HotpotQA denominator validity, external answer-performance validation, and failure-label reliability. Each component addresses a possible confound, but the paper's primary inferential unit remains the matched four-condition contrast.

The strongest supported claim is behavioral and condition-level: under the tested fixed evidence-availability interventions, the evaluated long-context and retrieve-then-read systems differ in how much recoverable evidence advantage survives evidence presentation, retrieval, reading, and output formatting. This claim is strongest for the reconstructed oracle-compatible datasets, local open-weight models, and retriever settings evaluated here. Scope boundaries are consolidated in Section~\ref{sec:limitations}.

\section{Limitations}
\label{sec:limitations}
The study measures observable condition-level behavior rather than mechanistic causal evidence use. ONCU measures how scores change when evidence availability is changed under a fixed protocol. It does not prove that a particular internal computation, attention path, or hidden state causally used a specific passage. Such claims would require additional interventions such as counterfactual passage replacement, activation patching, causal mediation, or mechanistic tracing.

The oracle-evidence condition is an isolated-evidence reference, not a perfect upper bound. Annotated evidence may omit helpful neighboring context, aliases, or redundant support, while retrieved chunks and full contexts can contain auxiliary information. Different oracle-evidence extraction policies could change the denominator-valid subset and the frequency of above-oracle or below-baseline raw ONCU behavior. For this reason, raw ONCU above 1 and below 0 are preserved as diagnostic regimes, and clipped ONCU is used only as a reporting convention for recovered-fraction summaries.

ONCU aggregation is conditional on denominator-valid groups satisfying $S_{\mathrm{oracle}}>S_{\mathrm{no}}$. This condition is broad in Controlled-ONCU-safe16K but narrower in HotpotQA-ONCU-200. The HotpotQA conclusions should therefore be read in two layers: ONCU supports claims over oracle-improving groups, while dataset-level conclusions rely on denominator-free answer/evidence scores, HotpotQA-500 robustness, retrieval-budget checks, and diagnostic summaries.

The reconstructed HotpotQA-ONCU and 2WikiMultiHopQA-ONCU datasets are designed as oracle-compatible diagnostic versions of the original benchmarks. They retain examples that can be represented with visible passage identifiers and oracle evidence passages. This construction enables the four-condition protocol, while favoring examples with cleaner evidence mappings and excluding some cases with ambiguous or redundant support.

The model panel covers five local open-weight models across Qwen, Gemma, Llama, and Mistral families, but it does not cover hosted proprietary frontier systems, larger open-weight variants such as 70B-scale models, learned retrievers, supervised multi-hop retrieval policies, or domain-specific long-context systems. The empirical conclusions should therefore be read as strongest for the evaluated local open-weight long-context and retrieve-then-read behavior under deterministic decoding and fixed structured-output contracts.

The automatic categorical labels should be interpreted as aggregate descriptive support rather than item-level causal labeling. The human audit shows moderate annotator agreement but substantially lower rule-versus-final-human agreement. We therefore use the labels to summarize broad diagnostic patterns and rely on continuous answer, evidence, retrieval, and ONCU metrics for primary claims.

The retriever-family sensitivity experiment addresses the most direct lexical-baseline concern by rerunning the complete four-condition protocol for dense@16 and hybrid@16 retrieved inputs. It is still not an exhaustive retriever benchmark. It does not cover learned retrievers, supervised multi-hop retrieval policies, retriever fine-tuning, or every top-$k$ budget under full ONCU references. The broader reader-facing validation and five-model cross-encoder reranking audit therefore remain auxiliary sensitivity analyses, and future work should extend the matched protocol to learned retrieval systems and more retrieval budgets.

Finally, the external BABILong-200 and RULER-lite-240 experiments are answer-performance validations, not ONCU benchmarks. They do not instantiate matched no-evidence and oracle-evidence reference conditions. Extending the full four-condition protocol to these and other long-context benchmarks is future work.

\section*{Data and Code Availability}
\label{sec:data_code_availability}
{\sloppy
The implementation, fixed configurations, processed-data builders, released processed inputs, frozen experiment summaries, confidence intervals, and reproduction instructions are available in the fixed public GitHub release: \url{https://github.com/Haizhoux0517/long_context_cue/releases/tag/v1.0.1-jair}. The release is tagged as \texttt{v1.0.1-jair} and corresponds to the artifact snapshot used for this submission. Source code is released under the MIT License. Documentation, figures, tables, README files, and supplementary materials are released under CC BY 4.0. Third-party datasets and benchmark resources retain their original licenses and terms of use, as documented in \texttt{DATA\_LICENSES.md}.

The release contains the protocol code, fixed run configurations, core processed JSONL inputs, frozen result artifacts, runtime and model-environment records, and scripts for recomputing the reported tables. The submitted snapshot was verified with \path{scripts/check_release_artifacts.py --strict-data --strict-clean}. Table~\ref{tab:repo_artifact_map} summarizes the main artifact groups; the complete reviewer-facing path map is provided in Appendix~\ref{app:full_artifact_map}. Runtime records include hardware and software metadata, deterministic inference controls, and unavailable-tool diagnostics; exact Ollama, model digest, and quantization entries are recorded when available on the reproduction host.

\begin{table}[!htbp]
\renewcommand{\arraystretch}{1.08}
\caption{Compressed Code and Data Artifact Map. The full path-level map appears in Appendix~\ref{app:full_artifact_map}.}
\label{tab:repo_artifact_map}
\centering
\scriptsize
\setlength{\tabcolsep}{3pt}
\begin{tabular}{p{0.25\columnwidth}p{0.30\columnwidth}p{0.34\columnwidth}}
\toprule
Artifact group & Main locations & Reviewer audit role \\
\midrule
Protocol, runners, and scoring
& \path{longcue/}; \path{scripts/validate_diagnostic_protocol.py}; \path{scripts/recompute_oncu.py}
& Reruns the fixed four-condition protocol and recomputes ONCU summaries. \\

Processed inputs and builders
& \path{data/processed/}; \path{scripts/build_*_cue.py}; \path{scripts/build_ruler_lite.py}
& Audits how ONCU-compatible and external validation inputs are materialized. \\

Configurations and frozen results
& \path{configs/}; \path{experiment_backups/}
& Maps reported experiment families to fixed run settings and archived outputs. \\

Reproduction and runtime records
& \path{README_REPRODUCE.md}; \path{ARTIFACT_MANIFEST.md}; \path{RUNTIME_REPRODUCIBILITY_RECORD.md}
& Documents commands, dependencies, environment metadata, checksums, and release verification. \\
\bottomrule
\end{tabular}
\end{table}

\noindent\textbf{Processed inputs.}
The processed evaluation files are JSONL records with explicit sample identifiers, questions, passage-annotated contexts, gold answers, and metadata fields. The ONCU-compatible datasets also include oracle-evidence identifiers, which are used to construct the oracle-evidence condition and to compute evidence-overlap metrics. BABILong-200 and RULER-lite-240 are kept separate because their current adapters do not provide the full no-evidence and oracle-evidence reference protocol required for ONCU. Original public datasets, including HotpotQA, 2WikiMultiHopQA, BABILong, and RULER-style benchmark resources, should be obtained from their official providers subject to their licenses and terms of use.

\noindent\textbf{Regeneration paths.}
The repository records exact commands and checksums in \path{README_REPRODUCE.md} and \path{ARTIFACT_MANIFEST.md}. The main paper reports audit locations rather than printing long checksums inline. The principal regeneration paths are:
\begin{itemize}
    \item \textbf{Runtime and model environment record:} generated by \path{scripts/export_runtime_record.py} as \path{RUNTIME_REPRODUCIBILITY_RECORD.md} and \path{runtime_reproducibility_record.json}; the submitted record captures the Runpod A40 reproduction host, deterministic inference controls, PyTorch/CUDA metadata, NVIDIA driver/CUDA information, and requested local model tags.
    \item \textbf{2WikiMultiHopQA-ONCU-500:} generated by \path{scripts/build_2wiki_cue.py} with the fixed seed reported in the artifact manifest.
    \item \textbf{Retriever-family ablation:} produced by \path{scripts/run_retriever_family_ablation.py} and archived under \path{experiment_backups/retriever_family_ablation_20260527/}.
    \item \textbf{Retriever-family ONCU sensitivity:} generated by \path{scripts/prepare_retriever_family_oncu_sensitivity.py}, run with \path{configs/retriever_family_oncu_sensitivity/*.yaml}, and archived under \path{experiment_backups/retriever_family_oncu_sensitivity_20260602/}.
    \item \textbf{Five-model cross-encoder reranking audit:} configured with \path{configs/rerank_sensitivity/}, run with \path{RUN_CE_RERANK_CONFIG_LIST.sh}, summarized by \path{scripts/summarize_ce_rerank_five_model.py}, and represented in the release by lightweight summary tables under \path{experiment_backups/rerank_sensitivity_20260602/five_model_ce_rerank_summary/}.
    \item \textbf{RULER-lite-240:} generated by \path{scripts/build_ruler_lite.py}, evaluated by \path{scripts/run_ruler_lite_external.py}, summarized by \path{scripts/summarize_ruler_lite_external.py}, and archived under \path{experiment_backups/ruler_lite_external_20260530_final/}. The generated input file is recreated by the builder before rerunning the audit rather than treated as a checked-in core input. The Qwen3-14B run uses the API-level thinking-output control described in Section~8.
    \item \textbf{Controlled scaling:} generated by \path{scripts/build_controlled_scaling_cue.py}, summarized by \path{scripts/summarize_controlled_scaling.py}, and archived under \path{experiment_backups/controlled_scaling_20260527/}. The generated 3,200-sample input is regenerated by the released builder when rerunning this auxiliary audit.
    \item \textbf{Failure-taxonomy human validation:} exported by \path{scripts/export_failure_taxonomy_audit.py}, summarized by \path{scripts/summarize_failure_taxonomy_audit.py}, and archived under \path{experiment_backups/failure_taxonomy_human_validation_20260530/}.
\end{itemize}

If anonymous review is required, the repository URL should be replaced by an anonymized archival link or anonymous repository snapshot during submission and restored after review. Publication metadata fields such as DOI, volume, article number, and associate editor should be left blank or suppressed in the review PDF until assigned by JAIR.
}

\section{Conclusion}
Long-context and retrieval-augmented evaluation can be treated as a diagnostic protocol-and-estimation problem. Final answer accuracy alone cannot determine whether an answer reflects no-evidence answerability, whether a retriever preserved the necessary evidence chain, or whether the reader converted available evidence into the requested answer. The proposed four-condition protocol---no evidence, full context, retrieved evidence, and oracle-evidence reference---makes those contrasts observable. ONCU then provides a baseline-adjusted, denominator-valid estimate of recovered oracle-reference advantage under the matched protocol.

Across the five tested local open-weight models, the same qualitative pattern appears. Controlled synthetic settings show that compact or isolated evidence can be usable even when the same evidence embedded in long inputs is not reliably converted into oracle-reference advantage. In the tested HotpotQA and 2WikiMultiHopQA reconstructions, the evaluated retrieve-then-read inputs can lose support that remains available in full context. Scaling, model-family extension, retrieval-family, reader-facing, external-validation, and human-audit analyses support this bounded interpretation while also clarifying its limits.

The framework complements answer, evidence, and retrieval metrics by adding a matched diagnostic protocol, denominator-validity regime, and aggregate diagnostic-pattern audit for separating no-evidence answerability, evidence availability, reader utilization, denominator validity, and output stability. Future work should extend the matched four-condition protocol to stronger frontier and larger open-weight models, learned multi-hop retrievers and rerankers, domain-specific long-context tasks, and mechanistic interventions that can test causal evidence use directly.

\clearpage
\printbibliography[heading=bibintoc,title={References}]

\clearpage
\appendix

\section{Supplementary Audit Material}
\label{app:supplementary_audits}
\suppressfloats[t]
The main text is intentionally restricted to the evidence chain needed for the JAIR submission argument: protocol, estimator validity, primary ONCU results, controlled and realistic robustness checks, model-family coverage, and the matched retriever-family ONCU sensitivity experiment. The appendices below preserve detailed audit material that is useful for reviewers but would otherwise make the main article read like an experiment report.

\subsection{Complete Reviewer-Facing Artifact Map}
\label{app:full_artifact_map}
Table~\ref{tab:full_repo_artifact_map} gives the path-level code and data map summarized in Table~\ref{tab:repo_artifact_map}.

\begingroup
\scriptsize
\renewcommand{\arraystretch}{1.10}
\setlength{\tabcolsep}{2.5pt}
\begin{longtable}{>{\raggedright\arraybackslash}p{0.16\textwidth}
                  >{\raggedright\arraybackslash}p{0.34\textwidth}
                  >{\raggedright\arraybackslash}p{0.11\textwidth}
                  >{\raggedright\arraybackslash}p{0.31\textwidth}}
\caption{Complete Reviewer-Facing Code and Data Artifact Map. The paths are aligned with the submitted repository layout. ``Released'' denotes files or directories present in the repository release.}
\label{tab:full_repo_artifact_map}\\
\toprule
Artifact group & Repository path(s) & Status & Reviewer audit role \\
\midrule
\endfirsthead
\caption[]{Complete Reviewer-Facing Code and Data Artifact Map (continued).}\\
\toprule
Artifact group & Repository path(s) & Status & Reviewer audit role \\
\midrule
\endhead
\midrule
\multicolumn{4}{r}{\footnotesize Continued on next page}\\
\endfoot
\bottomrule
\endlastfoot
Experiment runner
& \path{longcue/run_experiment.py}; \path{longcue/}
& Released
& Executes the fixed diagnostic protocol and writes per-sample predictions and metrics. \\

Protocol and ONCU computation
& \path{scripts/validate_diagnostic_protocol.py}; \path{scripts/recompute_oncu.py}; \path{longcue/evaluation/oncu.py}
& Released
& Validates fixed protocol fields and recomputes raw and clipped ONCU summaries from per-sample metrics. \\

Dataset builders and adapters
& \path{scripts/build_controlled_cue.py}; \path{scripts/build_controlled_scaling_cue.py}; \path{scripts/build_hotpotqa_cue.py}; \path{scripts/build_2wiki_cue.py}; \path{scripts/build_babilong_cue.py}; \path{scripts/build_ruler_lite.py}; \path{longcue/data/}
& Released
& Builds controlled, controlled-scaling, HotpotQA, 2WikiMultiHopQA, BABILong, and RULER-lite processed inputs. \\

Run configurations
& \path{configs/*_200_core_final.yaml}; \path{configs/hotpotqa_*_500_core_robust.yaml}; \path{configs/twowiki_*_500_core.yaml}; \path{configs/model_family_extension/*.yaml}; \path{configs/babilong_*_200_external.yaml}; \path{configs/ablations/retriever_family_*.yaml}; \path{configs/ablations/reader_facing_retfam_*.yaml}; \path{configs/retriever_family_oncu_sensitivity/*.yaml}; \path{configs/scaling/controlled_scaling_*.yaml}; \path{scripts/run_ruler_lite_external.py}
& Released
& Records dataset paths, model names, retrieval settings, deterministic decoding settings, ablation settings, output directories, and the RULER-lite external validation runner. \\

Released processed JSONL inputs
& \path{data/processed/controlled_oncu_200_safe16k.jsonl}; \path{data/processed/hotpotqa_cue_200.jsonl}; \path{data/processed/hotpotqa_cue_500.jsonl}; \path{data/processed/twowiki_cue_500.jsonl}; \path{data/processed/babilong_cue_200_external.jsonl}
& Released
& Provides materialized inputs for the core ONCU-compatible runs, HotpotQA robustness run, and BABILong external validation. \\

Generated auxiliary inputs
& \path{data/processed/controlled_scaling_3200.jsonl}; \path{data/processed/ruler_lite_240.jsonl}
& Generated
& Recreated by the released controlled-scaling and RULER-lite builder scripts before rerunning those auxiliary audits. \\

Frozen result artifacts
& \path{experiment_backups/sci200_final_3model_20260525/}; \path{experiment_backups/twowiki_500_validation_20260527/}; \path{experiment_backups/model_family_extension_20260601/}; \path{experiment_backups/hotpotqa_500_robustness_20260525/}; \path{experiment_backups/babilong_200_external_20260526/}; \path{experiment_backups/ruler_lite_external_20260530_final/}; \path{experiment_backups/retriever_family_ablation_20260527/}; \path{experiment_backups/retriever_family_oncu_sensitivity_20260602/}; \path{experiment_backups/reader_facing_retriever_family_20260530/}; \path{experiment_backups/controlled_scaling_20260527/}; \path{experiment_backups/failure_taxonomy_human_validation_20260530/}
& Released
& Stores result CSV files, exported tables, confidence intervals, model-family outputs, external-validation summaries, retrieval audits, controlled scaling summaries, and human-validation outputs. \\

Cross-encoder reranking audit
& \path{configs/rerank_sensitivity/}; \path{RUN_CE_RERANK_CONFIG_LIST.sh}; \path{scripts/summarize_ce_rerank_five_model.py}; \path{experiment_backups/rerank_sensitivity_20260602/five_model_ce_rerank_summary/}
& Released summary/config
& Provides the seven-setting, five-model cross-encoder reranking appendix audit over HotpotQA-ONCU-200 and 2WikiMultiHopQA-ONCU-500. \\

Confidence intervals, retrieval ablations, and diagnostic summaries
& \path{scripts/bootstrap_sci200_final_ci.py}; \path{scripts/bootstrap_hotpotqa500_robustness_ci.py}; \path{scripts/bootstrap_babilong200_external_ci.py}; \path{scripts/summarize_ruler_lite_external.py}; \path{scripts/summarize_sci200_failure_breakdown.py}; \path{scripts/recompute_twowiki500_tables.py}; \path{scripts/run_retriever_family_ablation.py}; \path{scripts/prepare_retriever_family_oncu_sensitivity.py}; \path{scripts/summarize_reader_facing_retriever_results.py}; \path{scripts/summarize_controlled_scaling.py}
& Released
& Regenerates confidence intervals, external-validation summaries, diagnostic summaries, 2Wiki tables, retrieval-family summaries, reader-facing validation summaries, and controlled scaling summaries. \\

Failure-taxonomy human validation
& \path{scripts/export_failure_taxonomy_audit.py}; \path{scripts/summarize_failure_taxonomy_audit.py}; \path{experiment_backups/failure_taxonomy_human_validation_20260530/}
& Released
& Stores the blind audit sample, annotation codebook, annotator files, adjudicated labels, agreement summaries, confusion matrices, and manifest. \\

Runtime and model environment record
& \path{scripts/export_runtime_record.py}; \path{RUNTIME_REPRODUCIBILITY_RECORD.md}; \path{runtime_reproducibility_record.json}
& Released
& Records deterministic inference controls, package versions, GPU/VRAM, PyTorch/CUDA, NVIDIA driver/CUDA information, model tags, context-window settings, and unavailable runtime commands. \\

Release audit checker
& \path{scripts/check_release_artifacts.py}
& Released
& Verifies declared repository artifacts, including core files, released processed inputs, frozen result directories, and auxiliary summary/config artifacts. \\

Reproduction and dependencies
& \path{README_REPRODUCE.md}; \path{ARTIFACT_MANIFEST.md}; \path{requirements.txt}; \path{pyproject.toml}
& Released
& Documents environment setup, model names, execution commands, expected outputs, dependency metadata, and the mapping from paper tables to repository artifacts. \\
\end{longtable}
\endgroup
\FloatBarrier

\subsection{Metric-Failure Counterexample Table}
\label{app:metric_failure_modes}
Table~\ref{tab:metric_failure_modes} preserves the detailed counterexample summary behind the joint-observability argument in Section~\ref{sec:theoretical_properties}.

\begin{center}
\renewcommand{\arraystretch}{1.08}
\captionof{table}{Formal Failure Modes Behind the Joint-Observability Proposition. Each metric captures a useful quantity but leaves at least one term of the recovered-evidence-advantage target unidentified.}
\label{tab:metric_failure_modes}
\scriptsize
\setlength{\tabcolsep}{2pt}
\begin{tabular}{p{0.18\textwidth}p{0.25\textwidth}p{0.25\textwidth}p{0.25\textwidth}}
\toprule
Metric & What it measures & Counterexample & Missing reference \\
\midrule
Full-context accuracy & Final answer correctness under long input & Same full-context score with very different no-evidence baselines & No-evidence baseline and oracle-reference denominator \\
Retrieval recall@$k$ & Pre-reader evidence availability & Gold passages retrieved but reader integrates them incorrectly & Reader-side answer conversion \\
Evidence F1 & Citation/evidence overlap & Correct cited passages but wrong final answer & Answer-side correctness \\
Oracle gap & Distance from isolated-evidence reference & Small gap caused by high no-evidence answerability & No-evidence adjustment \\
Context gain & Improvement over no-evidence baseline & Same gain but different recoverable oracle-reference advantage & Oracle-reference denominator \\
ONCU with audits & Conditional recovered oracle-reference advantage & Invalid if oracle evidence does not improve over no evidence & Denominator-free companion metrics for dataset-wide interpretation \\
\bottomrule
\end{tabular}
\end{center}

\subsection{Comparison with Alternative Diagnostic Scores}
\label{app:metric_comparison}
\label{subsec:metric_comparison}

ONCU is intended to complement, not replace, standard answer, evidence, and retrieval metrics. Raw answer F1 and exact match measure whether the final answer is correct, but they do not distinguish contextual evidence use from no-evidence answer priors or parametric knowledge. Evidence F1 measures whether cited passages overlap with oracle evidence, but a model can cite relevant evidence and still fail to integrate or convert it into the correct answer. Retrieval recall@$k$ measures whether the retriever exposes the required evidence before answer generation, but it does not measure whether the reader uses that evidence. Oracle-gap and context-gain scores each control one side of the diagnostic comparison, but not both. ONCU is designed for the narrower question of how much oracle-evidence advantage a contextual condition recovers after adjusting for what the same model can answer without evidence. Because ONCU filters non-positive denominators, the empirical sections also report unnormalized example-level answer and evidence scores; these raw-score analyses serve as denominator-free sensitivity checks on the direction of the reported effects.

Table~\ref{tab:metric_comparison_alternatives} summarizes these distinctions. The comparison is important because several failure patterns in this paper would be hard to interpret from any single raw metric. A high full-context answer F1 can partly reflect no-evidence answerability; high evidence overlap can coexist with answer-conversion failure; and high retrieval recall can still lead to low reader utilization. ONCU addresses a distinct diagnostic question by normalizing a contextual score between the no-evidence baseline and the oracle-evidence reference.

\begin{table}[!htbp]
\renewcommand{\arraystretch}{1.08}
\caption{Comparison of ONCU with Alternative Diagnostic Scores. Baseline indicates whether the score adjusts for no-evidence answerability, Oracle-ref. indicates whether it accounts for an isolated-evidence reference, and Evidence indicates whether the score directly evaluates evidence availability or overlap.}
\label{tab:metric_comparison_alternatives}
\centering
\scriptsize
\setlength{\tabcolsep}{2pt}
\begin{tabular}{p{0.17\textwidth}ccp{0.17\textwidth}p{0.20\textwidth}p{0.27\textwidth}}
\toprule
Metric & Baseline & Oracle-ref. & Evidence & Primary use & Main limitation \\
\midrule
Raw answer F1 / EM
& No
& No
& No
& Final answer correctness
& Can conflate context use with no-evidence priors or parametric knowledge. \\

Evidence F1
& No
& No
& Post-reader evidence overlap
& Cited evidence quality
& Can be high when the final answer is converted incorrectly. \\

Retrieval recall@$k$
& No
& No
& Pre-reader availability
& Retriever evidence coverage
& Does not measure whether the reader uses retrieved evidence. \\

Oracle gap $S_{\mathrm{oracle}}-S_c$
& No
& Partially
& Optional
& Distance from isolated-evidence reference
& Does not adjust for no-evidence answerability. \\

Context gain $S_c-S_{\mathrm{no}}$
& Yes
& No
& Optional
& Improvement over no-evidence baseline
& Does not indicate how much recoverable evidence advantage was captured. \\

ONCU raw
& Yes
& Yes
& Via chosen score field
& Recovered oracle-evidence advantage
& Requires a positive oracle-over-baseline denominator. \\

ONCU clipped
& Yes
& Yes
& Via chosen score field
& Aggregate recovered fraction
& Clips diagnostic extremes above 1 or below 0. \\
\bottomrule
\end{tabular}
\end{table}

Table~\ref{tab:metric_comparison_case_studies} gives representative cases from the released artifacts. The controlled examples show why raw answer F1 alone is insufficient: Qwen3-14B has non-trivial full-context F1 on the controlled safe16K setting, but ONCU shows that the full-context condition recovers only about half of the oracle-evidence advantage, whereas compact retrieved evidence nearly saturates the oracle reference. The realistic multi-hop examples show the opposite phenomenon: full context can preserve evidence-chain coverage better than deterministic retrieve-then-read input, so retrieval can reduce ONCU even when the retrieved context is shorter. The controlled 32K cases further show why position-aware ONCU is useful: early evidence can collapse to zero recovered advantage while the final evidence decile remains recoverable. Finally, retrieval-only rows illustrate that retrieval recall and chain coverage are pre-reader diagnostics; they identify evidence availability limitations but do not by themselves measure reader utilization.

\begin{table}[!htbp]
\renewcommand{\arraystretch}{1.06}
\caption{Metric-Comparison Case Studies. Raw answer F1, evidence F1 or chain coverage, context gain, oracle gap, and ONCU answer different diagnostic questions. Dashes indicate metrics that are not defined for retrieval-only rows or for aggregated controlled-scaling cells without evidence-overlap summaries.}
\label{tab:metric_comparison_case_studies}
\centering
\scriptsize
\setlength{\tabcolsep}{1.8pt}
\begin{tabular}{p{0.20\textwidth}p{0.10\textwidth}p{0.09\textwidth}rrrrrp{0.25\textwidth}}
\toprule
Case & Model/Retriever & Setting & Ans./Recall & Ev./Coverage & Gain & Gap & ONCU & Diagnostic reading \\
\midrule
Controlled full context: raw F1 hides oracle-referenced under-recovery
& Qwen3-14B
& Full context
& 0.526
& 0.706
& 0.518
& 0.469
& 0.535
& Full-context condition under-recovers oracle advantage. \\

Controlled retrieved evidence: compact evidence recovers oracle advantage
& Qwen3-14B
& Retrieved evidence
& 0.993
& 0.972
& 0.985
& 0.002
& 0.994
& Compact evidence recovers most oracle advantage. \\

HotpotQA full context: raw and ONCU agree that full context helps
& Qwen2.5-14B
& Full context
& 0.733
& 0.688
& 0.491
& 0.035
& 0.906
& Full context preserves multi-hop evidence better than retrieval. \\

HotpotQA retrieved evidence: retrieval coverage lowers ONCU
& Qwen2.5-14B
& Retrieved evidence
& 0.590
& 0.478
& 0.349
& 0.178
& 0.639
& Retrieved input improves over no evidence, but loses recoverable oracle advantage. \\

2Wiki full context: non-trivial raw F1 but moderate ONCU
& Qwen3-14B
& Full context
& 0.560
& 0.663
& 0.287
& 0.174
& 0.534
& Realistic multi-hop utilization remains incomplete. \\

2Wiki retrieved evidence: low ONCU despite compact input
& Qwen3-14B
& Retrieved evidence
& 0.400
& 0.352
& 0.126
& 0.335
& 0.367
& Retrieved evidence loses recoverable advantage. \\

Controlled 32K early evidence: raw context gain and ONCU collapse
& Qwen3-14B
& Full context
& 0.000
& --
& 0.000
& 0.996
& 0.000
& Early evidence is not converted into oracle-referenced gain. \\

Controlled 32K final-decile evidence: ONCU recovers at the end
& Qwen3-14B
& Full context
& 0.988
& --
& 0.988
& 0.000
& 1.000
& Final-decile evidence remains recoverable under full context. \\

Retrieval-only HotpotQA lexical@3: recall before reading
& Lexical
& top-$k=3$
& 0.633
& 0.540
& --
& --
& --
& Retrieval may limit multi-hop coverage before reading. \\

Retrieval-only 2Wiki dense@16: coverage increases with budget
& Dense
& top-$k=16$
& 0.842
& 0.756
& --
& --
& --
& Higher retrieval budget improves evidence-chain coverage but remains pre-reader. \\
\bottomrule
\end{tabular}
\end{table}

\subsection{Aggregate Raw-vs-Clipped ONCU Audit}
\label{app:raw_clipped_audit}

The main ONCU tables report clipped group-averaged ONCU because clipped values support the recovered-fraction interpretation. The raw-ratio audit in Table~\ref{tab:aggregate_raw_clipped_audit} makes visible when a contextual condition exceeds the oracle-evidence reference or falls below the no-evidence baseline before clipping. Raw ratios are computed from the group-averaged score columns in Table~\ref{tab:sci200_oncu_results} and are interpreted as clipping-sensitivity diagnostics rather than as replacements for the official group-averaged ONCU values.

\begin{table*}[!tbp]
\renewcommand{\arraystretch}{1.10}
\caption{Aggregate Raw-vs-Clipped ONCU Audit for the Final 200-Sample Matrix. Raw ratios are computed from the group-averaged score columns in Table~\ref{tab:sci200_oncu_results} as $(S_c-S_{\mathrm{no}})/(S_{\mathrm{oracle}}-S_{\mathrm{no}})$. They are used only as a clipping-sensitivity audit; the official ONCU results remain the group-averaged values reported in Table~\ref{tab:sci200_oncu_results}.}
\label{tab:aggregate_raw_clipped_audit}
\centering
\scriptsize
\setlength{\tabcolsep}{3pt}
\begin{tabular}{llcccp{0.30\textwidth}}
\toprule
Model & Dataset & Condition & Aggregate raw ratio & After clipping & Diagnostic reading \\
\midrule
Qwen2.5-14B & Controlled-safe16K-200 & Full context & 0.601 & 0.601 & In-range under-recovery of oracle advantage. \\
Qwen2.5-14B & Controlled-safe16K-200 & Retrieved evidence & 1.079 & 1.000 & Above-oracle aggregate behavior; compact retrieved context can exceed the isolated oracle reference. \\
Qwen3-14B & Controlled-safe16K-200 & Full context & 0.531 & 0.531 & In-range full-context under-recovery. \\
Qwen3-14B & Controlled-safe16K-200 & Retrieved evidence & 0.994 & 0.994 & Near-complete recovery without above-oracle aggregate behavior. \\
Gemma3-12B & Controlled-safe16K-200 & Full context & 0.517 & 0.517 & In-range full-context under-recovery. \\
Gemma3-12B & Controlled-safe16K-200 & Retrieved evidence & 0.845 & 0.845 & In-range recovered fraction. \\
Qwen2.5-14B & HotpotQA-ONCU-200 & Full context & 0.948 & 0.948 & In-range, high full-context recovery. \\
Qwen2.5-14B & HotpotQA-ONCU-200 & Retrieved evidence & 0.656 & 0.656 & In-range but below full context, consistent with retrieval-chain loss. \\
Qwen3-14B & HotpotQA-ONCU-200 & Full context & 0.820 & 0.820 & In-range, higher than retrieved evidence. \\
Qwen3-14B & HotpotQA-ONCU-200 & Retrieved evidence & 0.565 & 0.565 & In-range retrieved-evidence limitation. \\
Gemma3-12B & HotpotQA-ONCU-200 & Full context & 0.764 & 0.764 & In-range, higher than retrieved evidence. \\
Gemma3-12B & HotpotQA-ONCU-200 & Retrieved evidence & 0.586 & 0.586 & In-range retrieved-evidence limitation. \\
\bottomrule
\end{tabular}
\end{table*}

\subsection{Failure Diagnosis}
\label{app:failure_taxonomy_rules}
We diagnose failures according to the relationship between answer correctness and evidence correctness. Table~\ref{tab:failure_taxonomy_rules} summarizes the operational rules used to assign the main failure categories. The labels are intended as categorical diagnostic annotations rather than replacements for continuous answer F1 or evidence F1. In particular, the success rate in this taxonomy can be stricter than relaxed answer F1 because it depends jointly on answer and evidence behavior.

\begin{table}[!htbp]
\renewcommand{\arraystretch}{1.12}
\caption{Operational Rules for Failure-Type Assignment.}
\label{tab:failure_taxonomy_rules}
\centering
\footnotesize
\setlength{\tabcolsep}{3pt}
\begin{tabular}{p{0.28\columnwidth} p{0.62\columnwidth}}
\toprule
Label & Operational rule \\
\midrule
Localization
& The predicted evidence has zero or near-zero overlap with the oracle evidence and the answer is incorrect. \\
Selection
& The prediction cites some relevant evidence, but the cited set is incomplete, distractor-heavy, or misses a required supporting passage. \\
Integration
& The cited evidence substantially overlaps with the oracle evidence, but the final answer remains incorrect, indicating a failure to combine the evidence. \\
Conversion
& The cited evidence supports the correct answer, but the generated answer has the wrong value, entity, format, or normalized answer form. \\
Success
& The answer satisfies the answer-correctness criterion and the cited evidence satisfies the evidence-correctness criterion. \\
Parse
& The structured model output cannot be parsed into the required answer-and-evidence format. \\
\bottomrule
\end{tabular}
\end{table}

\subsection{Human Validation of Failure-Type Assignment}
\label{app:human_failure_validation}
\label{subsec:human_failure_validation}

The failure taxonomy in Table~\ref{tab:failure_taxonomy_rules} is rule-based and is therefore used as a scalable diagnostic approximation rather than as ground-truth causal attribution. To evaluate whether the operational labels align with human judgments, we conducted a blinded human-validation audit on a stratified sample of 300 failed predictions. The sample was drawn across datasets, models, evidence-availability conditions, and rule-assigned failure categories. Two annotators independently assigned one of six labels: localization, selection, integration, conversion, parse-format, or ambiguous. One annotator was the author and the other was an anonymous independent annotator who requested not to be named. During the initial annotation pass, annotators were blind to the rule-based labels and to each other's labels.

Before adjudication, the annotators agreed on 213 of 300 items, yielding raw agreement of 0.710 and Cohen's $\kappa=0.588$ \parencite{cohen1960coefficient}. The remaining 87 disagreements were resolved by blind adjudication to obtain final human labels. Against these adjudicated labels, the rule-based taxonomy matched 154 of 300 items, corresponding to raw agreement of 0.513 and Cohen's $\kappa=0.355$. Table~\ref{tab:failure_taxonomy_human_validation} summarizes the validation results, and Table~\ref{tab:failure_taxonomy_label_distribution} reports the label distributions.

The human-validation audit supports two interpretations. First, the annotators show moderate agreement on a difficult six-way failure-labeling problem, which indicates that the taxonomy captures recognizable failure modes instead of arbitrary categories. Second, the lower rule-versus-final-human agreement shows that the automatic rule labels should not be interpreted as item-level causal ground truth. We therefore use the automatic taxonomy for aggregate diagnostic analysis, preserve continuous answer and evidence metrics as the primary measurements, and release the human-validation labels, adjudication outcomes, agreement summaries, and confusion matrices for auditability.

\begin{table}[!htbp]
\renewcommand{\arraystretch}{1.10}
\caption{Human Validation of Failure-Type Assignment. Two annotators independently labeled a stratified audit sample of failed predictions. Disagreements were resolved by blind adjudication to produce final human labels.}
\label{tab:failure_taxonomy_human_validation}
\centering
\small
\setlength{\tabcolsep}{5pt}
\begin{tabular}{lr}
\toprule
Quantity & Value \\
\midrule
Audit items & 300 \\
Pre-adjudication agreed items & 213 \\
Pre-adjudication disagreements & 87 \\
Pre-adjudication raw agreement & 0.710 \\
Pre-adjudication Cohen's $\kappa$ & 0.588 \\
Rule vs. final-human raw agreement & 0.513 \\
Rule vs. final-human Cohen's $\kappa$ & 0.355 \\
\bottomrule
\end{tabular}
\end{table}

\begin{table}[!htbp]
\renewcommand{\arraystretch}{1.08}
\caption{Failure-Label Distributions in the Human-Validation Audit. Counts are over the same 300 audited failed predictions.}
\label{tab:failure_taxonomy_label_distribution}
\centering
\small
\setlength{\tabcolsep}{4pt}
\begin{tabular}{lrrrr}
\toprule
Label & Annotator A & Annotator B & Rule-Based & Final Human \\
\midrule
Localization & 71 & 78 & 74 & 81 \\
Selection & 56 & 75 & 71 & 84 \\
Integration & 30 & 18 & 74 & 16 \\
Conversion & 136 & 120 & 81 & 110 \\
Parse-format & 7 & 8 & 0 & 8 \\
Ambiguous & 0 & 1 & 0 & 1 \\
\bottomrule
\end{tabular}
\end{table}

\subsection{Final 200-Sample Bootstrap Confidence Intervals}
\label{app:bootstrap_ci}
Table~\ref{tab:sci200_oncu_ci_results} reports group-level bootstrap confidence intervals for ONCU-Relaxed-F1 in the final 200-sample matrix. The intervals support the two main diagnostic patterns. In the controlled setting, retrieved-evidence ONCU remains substantially higher than full-context ONCU for all three models: Qwen2.5-14B obtains 0.981 [0.955, 1.000] under retrieved evidence compared with 0.583 [0.506, 0.663] under full context; Qwen3-14B obtains 0.994 [0.985, 1.000] compared with 0.535 [0.451, 0.619]; and Gemma3-12B obtains 0.842 [0.782, 0.897] compared with 0.515 [0.438, 0.591]. These intervals reinforce the conclusion that the controlled benchmark exposes a full-context utilization gap instead of a general inability to answer from compact evidence.

The HotpotQA-derived setting shows the reverse pattern on denominator-valid groups. Full-context ONCU is consistently higher than retrieved-evidence ONCU: Qwen2.5-14B obtains 0.906 [0.843, 0.958] compared with 0.639 [0.511, 0.755], Qwen3-14B obtains 0.787 [0.679, 0.881] compared with 0.557 [0.441, 0.675], and Gemma3-12B obtains 0.719 [0.604, 0.819] compared with 0.536 [0.428, 0.645]. Together with the lower retrieved-evidence F1 values in Table~\ref{tab:sci200_main_results}, these intervals support the scoped interpretation that retrieved evidence can lose recoverable support in realistic multi-hop samples: sample-level answer/evidence metrics support the direction over evaluated examples, and ONCU supports it over oracle-improving groups.

\begin{table}[!htbp]
\renewcommand{\arraystretch}{1.12}
\caption{Bootstrap 95\% Confidence Intervals for Final 200-Sample ONCU-Relaxed-F1. ONCU is bootstrapped over valid metadata groups using 5,000 bootstrap resamples. Full denotes full-context input and Ret. denotes retrieved evidence.}
\label{tab:sci200_oncu_ci_results}
\centering
\scriptsize
\setlength{\tabcolsep}{3pt}
\begin{tabular}{llccc}
\toprule
Model & Dataset & Full ONCU [95\% CI] & Ret. ONCU [95\% CI] & Valid Groups \\
\midrule
Qwen2.5-14B & Controlled-safe16K-200 & 0.583 [0.506, 0.663] & \textbf{0.981 [0.955, 1.000]} & 134 \\
Qwen2.5-14B & HotpotQA-ONCU-200 & \textbf{0.906 [0.843, 0.958]} & 0.639 [0.511, 0.755] & 29 \\
Qwen3-14B & Controlled-safe16K-200 & 0.535 [0.451, 0.619] & \textbf{0.994 [0.985, 1.000]} & 133 \\
Qwen3-14B & HotpotQA-ONCU-200 & \textbf{0.787 [0.679, 0.881]} & 0.557 [0.441, 0.675] & 28 \\
Gemma3-12B & Controlled-safe16K-200 & 0.515 [0.438, 0.591] & \textbf{0.842 [0.782, 0.897]} & 133 \\
Gemma3-12B & HotpotQA-ONCU-200 & \textbf{0.719 [0.604, 0.819]} & 0.536 [0.428, 0.645] & 28 \\
\bottomrule
\end{tabular}
\end{table}

\subsection{Final 200-Sample Failure-Type Analysis}
\label{app:failure_breakdown}
Table~\ref{tab:failure_breakdown_final} reports the final 200-sample failure-type breakdown for the two contextual conditions. The taxonomy is used as a diagnostic annotation and should be interpreted together with continuous answer and evidence metrics. Its categorical success label is stricter than relaxed answer F1 because it depends jointly on answer and evidence behavior.

The controlled setting shows a consistent reduction in evidence-localization labels under retrieved evidence. For Qwen2.5-14B, localization labels decrease from 36.0\% under full-context input to 0.0\% under retrieved evidence. Qwen3-14B shows a similar decrease from 23.5\% to 0.0\%, and Gemma3-12B decreases from 10.5\% to 0.0\%. This is consistent with the interpretation that the controlled benchmark exposes a full-context evidence-utilization contrast: when the relevant evidence is compactly provided, localization-like labels are largely removed by the automatic taxonomy.

The HotpotQA-derived setting shows the opposite pattern. Full-context input has very low localization failure rates, ranging from 0.5\% to 1.5\% across models. In contrast, retrieved-evidence input increases localization failures to 23.0\% for Qwen2.5-14B, 22.5\% for Qwen3-14B, and 26.0\% for Gemma3-12B. This is consistent with the retrieval-coverage interpretation of the HotpotQA results: deterministic lexical retrieval can discard or truncate supporting facts required for multi-hop reasoning. The automatic labels are used as aggregate descriptive support, not as item-level causal ground truth.

The Gemma3-12B controlled run further illustrates the diagnostic value of the framework. Although retrieved evidence removes localization failures, Gemma3-12B retains higher integration failures than the Qwen-family models. Its full-context condition also contains a small number of structured-output parsing failures. The failure taxonomy complements ONCU by describing whether low utilization scores are associated, in aggregate, with missing evidence, evidence integration, answer conversion, or structured-output instability.

\begin{table}[!htbp]
\renewcommand{\arraystretch}{1.10}
\caption{Final 200-Sample Failure-Type Breakdown for Contextual Conditions. Rates are percentages over 200 examples per row. Loc., Sel., Int., Conv., Succ., and Parse denote evidence localization failure, evidence selection failure, evidence integration failure, answer conversion failure, categorical success, and structured-output parsing failure, respectively. The success label is stricter than relaxed answer F1 and is used only as a diagnostic category.}
\label{tab:failure_breakdown_final}
\centering
\scriptsize
\setlength{\tabcolsep}{2.5pt}
\begin{tabular}{lllrrrrrr}
\toprule
Model & Dataset & Condition & Loc. & Sel. & Int. & Conv. & Succ. & Parse \\
\midrule
Qwen2.5-14B & Controlled-safe16K-200 & Full Context & 36.0 & 0.5 & 10.0 & 38.5 & 15.0 & 0.0 \\
Qwen2.5-14B & Controlled-safe16K-200 & Retrieved Evidence & 0.0 & 3.0 & 3.0 & 43.5 & 50.5 & 0.0 \\
Qwen2.5-14B & HotpotQA-ONCU-200 & Full Context & 1.5 & 5.0 & 22.5 & 11.5 & 59.5 & 0.0 \\
Qwen2.5-14B & HotpotQA-ONCU-200 & Retrieved Evidence & 23.0 & 2.5 & 23.5 & 3.0 & 48.0 & 0.0 \\
\midrule
Qwen3-14B & Controlled-safe16K-200 & Full Context & 23.5 & 0.5 & 14.5 & 30.0 & 31.5 & 0.0 \\
Qwen3-14B & Controlled-safe16K-200 & Retrieved Evidence & 0.0 & 1.0 & 4.5 & 38.0 & 56.5 & 0.0 \\
Qwen3-14B & HotpotQA-ONCU-200 & Full Context & 0.5 & 7.0 & 28.0 & 9.5 & 55.0 & 0.0 \\
Qwen3-14B & HotpotQA-ONCU-200 & Retrieved Evidence & 22.5 & 3.5 & 27.0 & 4.0 & 43.0 & 0.0 \\
\midrule
Gemma3-12B & Controlled-safe16K-200 & Full Context & 10.5 & 0.0 & 52.5 & 25.0 & 9.0 & 3.0 \\
Gemma3-12B & Controlled-safe16K-200 & Retrieved Evidence & 0.0 & 7.0 & 32.5 & 19.5 & 41.0 & 0.0 \\
Gemma3-12B & HotpotQA-ONCU-200 & Full Context & 0.5 & 8.5 & 31.5 & 3.5 & 56.0 & 0.0 \\
Gemma3-12B & HotpotQA-ONCU-200 & Retrieved Evidence & 26.0 & 2.5 & 26.0 & 1.0 & 44.5 & 0.0 \\
\bottomrule
\end{tabular}
\end{table}

\subsection{Statistical Modeling of Diagnostic Effects}
\label{app:statistical_checks}
\label{subsec:statistical_modeling}

The preceding tables are descriptive summaries over fixed diagnostic conditions. To ensure that the main conclusions are not driven by unpaired table comparisons, we add a statistical support layer that respects the repeated-measures structure of the evaluation. The analysis is not used to select prompts, models, datasets, or retrieval settings; it is a post-hoc audit of the fixed released artifacts.

Table~\ref{tab:statistical_effects_support} reports paired effect-size estimates for the central claims. In the controlled 200-sample matrix, retrieved evidence outperforms full-context input for all three models, with paired relaxed-F1 gains of 0.331 for Gemma3-12B, 0.438 for Qwen2.5-14B, and 0.467 for Qwen3-14B. In HotpotQA-ONCU-200 and 2WikiMultiHopQA-ONCU-500, the sign reverses: full-context input outperforms retrieved evidence, supporting the interpretation that realistic multi-hop retrieved-evidence gaps often reflect retrieval coverage rather than reader-side conversion alone. The same table also quantifies the controlled scaling effects. Across all three models, the 4K--32K full-context ONCU drop is large, and the 32K final-decile advantage over early and middle deciles remains large after paired resampling.

\begin{table}[!htbp]
\renewcommand{\arraystretch}{1.08}
\caption{Statistical Support for Main Diagnostic Effects. Estimates are paired mean differences unless otherwise noted. Confidence intervals are paired bootstrap 95\% intervals. $p_{\mathrm{Holm}}$ controls family-wise error within the stated analysis family.}
\label{tab:statistical_effects_support}
\centering
\scriptsize
\setlength{\tabcolsep}{2.2pt}
\begin{tabular}{p{0.19\textwidth} p{0.16\textwidth} p{0.18\textwidth} r r r r}
\toprule
Analysis & Model & Contrast & Est. & 95\% CI & Effect & $p_{\mathrm{Holm}}$ \\
\midrule
Controlled-safe16K-200: retrieved minus full & Gemma3-12B & retrieved - full & 0.331 & [0.266, 0.393] & 0.712 & $<$.001 \\
Controlled-safe16K-200: retrieved minus full & Qwen2.5-14B & retrieved - full & 0.438 & [0.371, 0.503] & 0.920 & $<$.001 \\
Controlled-safe16K-200: retrieved minus full & Qwen3-14B & retrieved - full & 0.467 & [0.403, 0.530] & 0.965 & $<$.001 \\
HotpotQA-ONCU-200: full minus retrieved & Gemma3-12B & full - retrieved & 0.107 & [0.046, 0.165] & 0.249 & $<$.001 \\
HotpotQA-ONCU-200: full minus retrieved & Qwen2.5-14B & full - retrieved & 0.143 & [0.090, 0.202] & 0.343 & $<$.001 \\
HotpotQA-ONCU-200: full minus retrieved & Qwen3-14B & full - retrieved & 0.131 & [0.074, 0.186] & 0.318 & $<$.001 \\
2WikiMultiHopQA-ONCU-500: full minus retrieved & Gemma3-12B & full - retrieved & 0.142 & [0.093, 0.190] & 0.265 & $<$.001 \\
2WikiMultiHopQA-ONCU-500: full minus retrieved & Qwen2.5-14B & full - retrieved & 0.171 & [0.130, 0.210] & 0.370 & $<$.001 \\
2WikiMultiHopQA-ONCU-500: full minus retrieved & Qwen3-14B & full - retrieved & 0.160 & [0.122, 0.201] & 0.356 & $<$.001 \\
Scaling: 4K minus 32K full-context ONCU & Gemma3-12B & 4K - 32K full-context ONCU & 0.634 & [0.472, 0.759] & 2.501 & $<$.001 \\
Scaling: 32K final decile minus early/middle deciles & Gemma3-12B & pos\_09 - mean(pos\_00..pos\_07) at 32K & 0.704 & [0.679, 0.726] & 19.459 & $<$.001 \\
Scaling: 32K retrieved minus full-context ONCU & Gemma3-12B & retrieved - full at 32K & 0.692 & [0.537, 0.808] & 2.888 & $<$.001 \\
Scaling: 4K minus 32K full-context ONCU & Qwen2.5-14B & 4K - 32K full-context ONCU & 0.836 & [0.638, 0.980] & 2.608 & $<$.001 \\
Scaling: 32K final decile minus early/middle deciles & Qwen2.5-14B & pos\_09 - mean(pos\_00..pos\_07) at 32K & 0.975 & [0.961, 0.989] & 44.659 & $<$.001 \\
Scaling: 32K retrieved minus full-context ONCU & Qwen2.5-14B & retrieved - full at 32K & 0.837 & [0.638, 0.982] & 2.606 & $<$.001 \\
Scaling: 4K minus 32K full-context ONCU & Qwen3-14B & 4K - 32K full-context ONCU & 0.829 & [0.625, 0.985] & 2.455 & $<$.001 \\
Scaling: 32K final decile minus early/middle deciles & Qwen3-14B & pos\_09 - mean(pos\_00..pos\_07) at 32K & 1.000 & [1.000, 1.000] & -- & $<$.001 \\
Scaling: 32K retrieved minus full-context ONCU & Qwen3-14B & retrieved - full at 32K & 0.850 & [0.649, 1.000] & 2.514 & $<$.001 \\
\bottomrule
\end{tabular}
\end{table}

Table~\ref{tab:statistical_regression_support} reports regression-style checks. The controlled scaling regressions are fit over aggregated length--position ONCU cells. Their purpose is to test whether the length and position patterns remain visible when summarized as continuous effects rather than as heatmaps. The log-context-length coefficient is negative for all three models, while the position-fraction coefficient is positive, indicating lower ONCU at longer contexts and stronger utilization near later evidence positions. The retriever-family regression supports the retrieval-ablation interpretation: larger retrieval budgets increase full-chain coverage, while retriever-family effects differ by dataset. These regressions are diagnostic support for effect direction and magnitude; they do not replace the paired contrasts or the released per-sample and per-group metrics.

\begin{table}[!htbp]
\renewcommand{\arraystretch}{1.08}
\caption{Regression-Style Statistical Checks. Controlled scaling rows are OLS diagnostics over aggregated length--position ONCU cells. Retriever rows are OLS diagnostics over retrieval-summary cells. These models support effect direction and magnitude rather than replacing the paired analyses.}
\label{tab:statistical_regression_support}
\centering
\scriptsize
\setlength{\tabcolsep}{3pt}
\begin{tabular}{p{0.24\textwidth} p{0.18\textwidth} p{0.22\textwidth} r r r}
\toprule
Analysis & Model/Dataset & Term & Est. & SE & $p_{\mathrm{BH}}$ \\
\midrule
Scaling OLS & Gemma3-12B / Controlled-scaling-3200 & log2\_context\_length\_centered & -0.214 & 0.023 & $<$.001 \\
Scaling OLS & Gemma3-12B / Controlled-scaling-3200 & position\_fraction\_centered & 0.256 & 0.044 & $<$.001 \\
Scaling OLS & Gemma3-12B / Controlled-scaling-3200 & distance\_from\_end\_centered & -0.256 & 0.044 & $<$.001 \\
Scaling OLS & Qwen2.5-14B / Controlled-scaling-3200 & log2\_context\_length\_centered & -0.281 & 0.033 & $<$.001 \\
Scaling OLS & Qwen2.5-14B / Controlled-scaling-3200 & position\_fraction\_centered & 0.373 & 0.064 & $<$.001 \\
Scaling OLS & Qwen2.5-14B / Controlled-scaling-3200 & distance\_from\_end\_centered & -0.373 & 0.064 & $<$.001 \\
Scaling OLS & Qwen3-14B / Controlled-scaling-3200 & log2\_context\_length\_centered & -0.281 & 0.033 & $<$.001 \\
Scaling OLS & Qwen3-14B / Controlled-scaling-3200 & position\_fraction\_centered & 0.380 & 0.064 & $<$.001 \\
Scaling OLS & Qwen3-14B / Controlled-scaling-3200 & distance\_from\_end\_centered & -0.380 & 0.064 & $<$.001 \\
Retriever OLS & retrieval-only / HotpotQA-ONCU-200 + 2WikiMultiHopQA-ONCU-500 & log2\_top\_k\_centered & 0.101 & 0.014 & $<$.001 \\
Retriever OLS & retrieval-only / HotpotQA-ONCU-200 + 2WikiMultiHopQA-ONCU-500 & retriever\_dense\_vs\_lexical & 0.037 & 0.040 & 0.371 \\
Retriever OLS & retrieval-only / HotpotQA-ONCU-200 + 2WikiMultiHopQA-ONCU-500 & retriever\_hybrid\_vs\_lexical & 0.014 & 0.040 & 0.733 \\
Retriever OLS & retrieval-only / HotpotQA-ONCU-200 + 2WikiMultiHopQA-ONCU-500 & retriever\_oracle\_vs\_lexical & 0.427 & 0.040 & $<$.001 \\
\bottomrule
\end{tabular}
\end{table}

Together, the paired contrasts, adjusted significance diagnostics, and regression-style checks strengthen the main interpretation without changing its scope. Controlled long-context failures appear as statistically large full-context recovery gaps under the tested protocol. Realistic multi-hop retrieve-then-read failures are statistically consistent with retrieval-coverage limitations. The controlled scaling extension is not merely a visual heatmap pattern: length and evidence position are associated with systematic variation in oracle-referenced recovery across all three evaluated models.

\subsection{HotpotQA Retrieval-Budget Sensitivity}
\label{app:retrieval_budget}
The lower retrieved-evidence performance on HotpotQA-ONCU-200 can reflect retrieval budget as well as reader-side utilization. To separate these factors, we run a retrieval-budget sensitivity analysis for Qwen2.5-14B and Qwen3-14B on HotpotQA-ONCU-200, varying the deterministic lexical retrieval budget from top-$k=3$ to top-$k=5$ and top-$k=8$ while keeping the dataset, output contract, decoding policy, and evaluation pipeline fixed. The top-$k=3$ setting corresponds to the main final-matrix configuration.

Table~\ref{tab:hotpotqa_topk_ablation_qwen} shows that increasing the lexical retrieval budget partially improves retrieved-evidence performance, but does not close the gap to full-context ONCU. For Qwen2.5-14B, moving from top-$k=3$ to top-$k=8$ increases retrieved-evidence F1 from 0.478 to 0.520 and retrieved-evidence ONCU from 0.639 to 0.723. For Qwen3-14B, the same change increases retrieved-evidence F1 from 0.469 to 0.530 and retrieved-evidence ONCU from 0.557 to 0.634. These improvements indicate that the HotpotQA retrieved-evidence gap is partly related to evidence coverage.

However, the retrieved-evidence condition remains below the corresponding full-context ONCU values from Table~\ref{tab:sci200_oncu_results}. For Qwen2.5-14B, top-$k=8$ retrieved-evidence ONCU is 0.723, compared with full-context ONCU of 0.906. For Qwen3-14B, top-$k=8$ retrieved-evidence ONCU is 0.634, compared with full-context ONCU of 0.787. Increasing the lexical top-$k$ budget mitigates but does not eliminate the retrieved-evidence gap. Realistic multi-hop retrieval therefore requires better evidence ranking, multi-hop coverage, and distractor control rather than simply retrieving more lexical chunks.

\begin{table}[!htbp]
\renewcommand{\arraystretch}{1.12}
\caption{HotpotQA Retrieval-Budget Sensitivity for Qwen2.5-14B and Qwen3-14B. The table reports retrieved-evidence condition performance on HotpotQA-ONCU-200 under different lexical top-$k$ budgets. ONCU is computed with relaxed answer F1.}
\label{tab:hotpotqa_topk_ablation_qwen}
\centering
\scriptsize
\setlength{\tabcolsep}{5pt}
\begin{tabular}{llccccc}
\toprule
Model & Setting & Rel. F1 & Strict F1 & Ev. F1 & Ret. ONCU & Parse Err. \\
\midrule
Qwen2.5-14B & top-$k=3$ & 0.590 & 0.595 & 0.478 & 0.639 & 0 \\
Qwen2.5-14B & top-$k=5$ & 0.588 & 0.593 & 0.506 & 0.635 & 1 \\
Qwen2.5-14B & top-$k=8$ & 0.612 & 0.618 & 0.520 & 0.723 & 0 \\
\midrule
Qwen3-14B & top-$k=3$ & 0.558 & 0.562 & 0.469 & 0.557 & 0 \\
Qwen3-14B & top-$k=5$ & 0.602 & 0.606 & 0.515 & 0.636 & 0 \\
Qwen3-14B & top-$k=8$ & 0.609 & 0.613 & 0.530 & 0.634 & 0 \\
\bottomrule
\end{tabular}
\end{table}

The ablation therefore refines the interpretation of the HotpotQA results. Both models improve when the retrieval budget is expanded, showing that retrieval depth is a real protocol variable. At the same time, the improvement is incomplete: the top-$k=8$ retrieved-evidence condition still falls short of full-context ONCU. The results point to a broader multi-hop retrieval problem under this protocol: the retriever must recover both supporting facts, rank them sufficiently high, and avoid adding distracting evidence that complicates downstream reasoning. The full-context condition remains stronger in this setting because it preserves all candidate supporting facts, even though it places a larger utilization burden on the model.

\subsection{Retriever-Family Retrieval Ablation}
\label{app:retrieval_ablation}
\label{subsec:retriever_family_ablation}

The preceding HotpotQA top-$k$ analysis varies only the lexical retrieval budget. The next protocol variable is retriever family. We therefore run a retrieval-only ablation on HotpotQA-ONCU-200 and 2WikiMultiHopQA-ONCU-500 without changing the materialized examples, passage identifiers, chunk size, overlap, or evidence-overlap metrics.

The evaluated retrieval families are lexical retrieval, dense sentence-embedding retrieval, hybrid lexical--dense retrieval using Reciprocal Rank Fusion, a deterministic iterative query-expansion baseline, and oracle retrieval. The dense retriever uses a fixed off-the-shelf sentence-embedding model and is not trained on either benchmark. The hybrid retriever combines lexical and dense rankings using reciprocal ranks, avoiding score-scale calibration between sparse and dense retrieval. The iterative retriever is a deterministic expansion baseline instead of a learned multi-hop retriever: it retrieves initial passages from the question, expands the retrieval query with selected retrieved text, and merges the ranked outputs. The oracle condition is not a deployable retriever; it is included as an evidence-chain coverage reference.

Table~\ref{tab:retriever_family_ablation} reports the compact main-paper view of the ablation for top-$k=3$ and top-$k=16$. The complete top-$k=3,5,8,16$ retrieval-only outputs are released as CSV artifacts. On HotpotQA-ONCU-200, lexical retrieval is competitive at small retrieval budgets: at top-$k=3$, lexical retrieval obtains full-chain coverage of 0.540, compared with 0.405 for dense retrieval, 0.455 for hybrid retrieval, and 0.430 for deterministic iterative retrieval. At the larger top-$k=16$ budget, hybrid retrieval obtains the highest non-oracle full-chain coverage, 0.805, but also exposes the reader to a substantially higher distractor rate than the small-budget settings.

The 2WikiMultiHopQA-ONCU-500 pattern is different. Dense retrieval substantially improves full-chain coverage relative to lexical retrieval: dense top-$k=3$ obtains full-chain coverage of 0.472 compared with 0.368 for lexical top-$k=3$, and dense top-$k=16$ obtains 0.756 compared with 0.524 for lexical top-$k=16$. However, this coverage improvement comes with increased distractor exposure: dense top-$k=16$ has a distractor identifier rate of 0.511, compared with 0.260 for lexical top-$k=16$. The oracle rows confirm that the evidence mappings themselves are not the limiting factor; oracle retrieval reaches complete or near-complete full-chain coverage once enough oracle passages are permitted. Realistic retrieved-evidence gaps therefore reflect a trade-off among evidence-chain coverage, ranking quality, retrieval budget, and distractor exposure, instead of a single limitation of lexical retrieval.

\begin{table}[!htbp]
\renewcommand{\arraystretch}{1.08}
\caption{Retriever-Family Retrieval-Only Ablation on HotpotQA-ONCU-200 and 2WikiMultiHopQA-ONCU-500. Recall, F1, and full-chain coverage are computed against oracle evidence identifiers. Distr. denotes the distractor identifier rate among retrieved passages. Avg. Pass. is the average number of unique retrieved passages after chunk-to-passage aggregation.}
\label{tab:retriever_family_ablation}
\centering
\scriptsize
\setlength{\tabcolsep}{3pt}
\begin{tabular}{llrrrrrr}
\toprule
Dataset & Retriever & Top-$k$ & Recall & Chain Cov. & Ret. F1 & Distr. & Avg. Pass. \\
\midrule
HotpotQA-200 & Lexical & 3 & 0.633 & 0.540 & 0.640 & 0.085 & 1.89 \\
HotpotQA-200 & Lexical & 16 & 0.819 & 0.750 & 0.557 & 0.459 & 5.09 \\
HotpotQA-200 & Dense & 3 & 0.502 & 0.405 & 0.461 & 0.310 & 2.33 \\
HotpotQA-200 & Dense & 16 & 0.832 & 0.745 & 0.497 & 0.588 & 6.04 \\
HotpotQA-200 & Hybrid & 3 & 0.565 & 0.455 & 0.566 & 0.099 & 1.77 \\
HotpotQA-200 & Hybrid & 16 & 0.873 & 0.805 & 0.574 & 0.486 & 5.53 \\
HotpotQA-200 & Iterative & 3 & 0.548 & 0.430 & 0.565 & 0.084 & 1.67 \\
HotpotQA-200 & Iterative & 16 & 0.781 & 0.700 & 0.517 & 0.499 & 5.15 \\
HotpotQA-200 & Oracle & 3 & 0.963 & 0.865 & 0.979 & 0.000 & 2.36 \\
HotpotQA-200 & Oracle & 16 & 1.000 & 1.000 & 1.000 & 0.000 & 2.52 \\
\midrule
2Wiki-500 & Lexical & 3 & 0.468 & 0.368 & 0.487 & 0.033 & 1.28 \\
2Wiki-500 & Lexical & 16 & 0.623 & 0.524 & 0.525 & 0.260 & 3.02 \\
2Wiki-500 & Dense & 3 & 0.602 & 0.472 & 0.545 & 0.224 & 2.82 \\
2Wiki-500 & Dense & 16 & 0.842 & 0.756 & 0.548 & 0.511 & 5.79 \\
2Wiki-500 & Hybrid & 3 & 0.484 & 0.386 & 0.502 & 0.066 & 1.44 \\
2Wiki-500 & Hybrid & 16 & 0.726 & 0.628 & 0.570 & 0.331 & 3.95 \\
2Wiki-500 & Iterative & 3 & 0.230 & 0.160 & 0.246 & 0.044 & 0.69 \\
2Wiki-500 & Iterative & 16 & 0.526 & 0.418 & 0.429 & 0.295 & 2.90 \\
2Wiki-500 & Oracle & 3 & 0.940 & 0.762 & 0.966 & 0.000 & 2.24 \\
2Wiki-500 & Oracle & 16 & 1.000 & 1.000 & 1.000 & 0.000 & 2.48 \\
\bottomrule
\end{tabular}
\end{table}

The ablation refines the interpretation of the earlier retrieved-evidence results. On HotpotQA, the main lexical retriever is a competitive small-budget diagnostic condition: it is stronger than dense and hybrid retrieval at the main top-$k=3$ setting and remains competitive at top-$k=8$. On 2WikiMultiHopQA, dense retrieval is clearly stronger than lexical retrieval in full-chain coverage, showing that retriever family matters. Across both datasets, increasing top-$k$ improves recall and full-chain coverage but also increases distractor exposure. The main retrieved-evidence gap should be interpreted as a dataset-dependent evidence-chain retrieval problem rather than as a single failure mode of deterministic lexical retrieval.

\subsection{Reader-Facing Retriever-Family Validation}
\label{app:reader_facing_validation}
\label{subsec:reader_facing_retriever_family_validation}

The matched sensitivity experiment tests dense@16 and hybrid@16 under the full four-condition protocol. We additionally retain the broader reader-facing validation because it answers a different question: across lexical, dense, and hybrid retrieval and across top-$k \in \{3,8,16\}$, which retrieved input actually works best for the downstream reader? This validation passes retrieved contexts to Qwen2.5-14B, Qwen3-14B, and Gemma3-12B under the same answer contract and deterministic decoding policy. It covers HotpotQA-ONCU-200 and 2WikiMultiHopQA-ONCU-500 and contains 18{,}900 reader predictions.

This validation is reported with answer/evidence metrics rather than as a full ONCU matrix because it varies both retriever family and retrieval budget. The matched ONCU sensitivity above supplies the complete four-condition references for dense@16 and hybrid@16; this broader sweep supplies budget- and family-level reader-facing context for interpreting those results. We report answer F1, evidence F1, parse failures, and the corresponding retrieval-only chain-coverage and distractor diagnostics.

Table~\ref{tab:reader_facing_retfam_validation} summarizes the best reader-facing retrieved setting for each model--dataset pair and compares it with the main lexical top-$k=3$ retrieved setting. The pattern is not a uniform victory for a single retriever. On HotpotQA-ONCU-200, the best reader-facing configuration is lexical top-$k=16$ for Gemma3-12B, hybrid top-$k=16$ for Qwen2.5-14B, and dense top-$k=16$ for Qwen3-14B. On 2WikiMultiHopQA-ONCU-500, dense retrieval is the best setting for Gemma3-12B and Qwen2.5-14B, while hybrid retrieval is best for Qwen3-14B. In all six model--dataset pairs, the best setting uses top-$k=16$, but the larger budget also increases distractor exposure. Retrieval-family improvements can transfer to reader-side answer performance, but the transfer is conditional on the dataset, model, retrieval family, and budget.

\begin{table}[!htbp]
\renewcommand{\arraystretch}{1.10}
\caption{Reader-Facing Retriever-Family Validation. Lexical@3 is the main retrieved-evidence setting used in the core protocol. Best setting is selected by relaxed answer F1 among lexical, dense, and hybrid retrieval at top-$k \in \{3,8,16\}$. Chain coverage and distractor rate are retrieval-only diagnostics for the selected best setting. This auxiliary validation is reported as a budget-level reader-facing sweep rather than as a complete ONCU matrix across all budgets.}
\label{tab:reader_facing_retfam_validation}
\centering
\scriptsize
\setlength{\tabcolsep}{3pt}
\begin{tabular}{llcccrrrr}
\toprule
Dataset & Model & Lexical@3 F1 & Best Ret. & $k$ & Best F1 & Best Ev.F1 & Chain Cov. & Distr. \\
\midrule
HotpotQA-200 & Gemma3-12B & 0.552 & Lexical & 16 & 0.622 & 0.505 & 0.750 & 0.459 \\
HotpotQA-200 & Qwen2.5-14B & 0.572 & Hybrid & 16 & 0.663 & 0.566 & 0.805 & 0.486 \\
HotpotQA-200 & Qwen3-14B & 0.559 & Dense & 16 & 0.623 & 0.562 & 0.745 & 0.588 \\
2Wiki-500 & Gemma3-12B & 0.393 & Dense & 16 & 0.499 & 0.515 & 0.756 & 0.511 \\
2Wiki-500 & Qwen2.5-14B & 0.370 & Dense & 16 & 0.473 & 0.513 & 0.756 & 0.511 \\
2Wiki-500 & Qwen3-14B & 0.400 & Hybrid & 16 & 0.484 & 0.475 & 0.628 & 0.331 \\
\bottomrule
\end{tabular}
\end{table}

The reader-facing validation changes the paper's retrieval claim in an important way. It supports the view that retrieval-only coverage is an incomplete proxy for retrieve-then-read performance: dense retrieval is clearly useful on 2WikiMultiHopQA, but HotpotQA remains model-dependent, and the largest answer gains appear only when the retrieval budget is expanded. The result strengthens the diagnostic framing rather than producing a simple recommendation to replace lexical retrieval with dense retrieval. A retrieval system should be audited at both stages: whether it exposes the full evidence chain before reading, and whether the reader converts the exposed evidence into a correct answer.

\subsection{Cross-Encoder Reranking Audit}
\label{app:ce_reranking_audit}

To test whether the retrieved-context conclusion depends on the absence of a reranking stage, we add a two-stage cross-encoder reranking audit for all five evaluated local models. The audit uses hybrid lexical--dense retrieval as a first-stage candidate generator, followed by \texttt{cross-encoder/ms-marco-MiniLM-L6-v2} to rescore query--chunk pairs. We instantiate seven reranked retrieved-context variants: CE@32$\rightarrow\{8,16\}$, CE@64$\rightarrow\{8,16,24\}$, and CE@128$\rightarrow\{16,24\}$, where the first number is the first-stage candidate pool and the second is the final reader budget. For this reader-facing reranking audit, ONCU is computed by joining each reranked retrieved-condition prediction with the existing no-evidence and oracle-evidence references by sample identifier; the values are interpreted as sample-level sensitivity diagnostics rather than replacements for the metadata-group ONCU tables used in the main four-condition matrix.

Table~\ref{tab:five_model_ce_rerank_audit} reports the best answer-F1 reranked setting for each model--dataset pair. Reranking narrows retrieved-context gaps in several rows, but it does not reduce retrieval-augmented behavior to a single best setting. On HotpotQA-200, the selected rows vary across CE@64 and CE@128 candidate pools depending on the model. On 2Wiki-500, all five best answer-F1 rows use CE@128$\rightarrow$24. The ONCU-best rows are not always identical to the answer-F1-best rows, reinforcing the diagnostic distinction among answer conversion, evidence overlap, and oracle-referenced recovery.

\begin{table*}[!tbp]
\renewcommand{\arraystretch}{1.08}
\caption{Five-Model Cross-Encoder Reranking Audit. CE@$m\rightarrow k$ denotes hybrid first-stage retrieval with $m$ candidates, cross-encoder reranking, and final reader budget $k$. The table reports the best answer-F1 reranked setting for each model--dataset pair; ONCU is computed by joining each reranked retrieved prediction with the corresponding no-evidence and oracle-evidence references by sample identifier.}
\label{tab:five_model_ce_rerank_audit}
\centering
\scriptsize
\setlength{\tabcolsep}{3pt}
\begin{tabular}{lllrrrrr}
\toprule
Dataset & Model & Best reranked setting & Answer F1 & Evidence F1 & Valid denom. & ONCU & Parse err. \\
\midrule
HotpotQA-200 & Gemma3-12B & CE@128$\rightarrow$24 & 0.665 & 0.549 & 121/200 & 0.771 & 0 \\
HotpotQA-200 & Llama3.1-8B & CE@128$\rightarrow$24 & 0.657 & 0.538 & 119/200 & 0.716 & 0 \\
HotpotQA-200 & Mistral-Small3.1-24B & CE@64$\rightarrow$24 & 0.735 & 0.649 & 116/200 & 0.788 & 0 \\
HotpotQA-200 & Qwen2.5-14B & CE@64$\rightarrow$16 & 0.674 & 0.574 & 121/200 & 0.773 & 0 \\
HotpotQA-200 & Qwen3-14B & CE@64$\rightarrow$16 & 0.637 & 0.561 & 124/200 & 0.724 & 1 \\
\midrule
2Wiki-500 & Gemma3-12B & CE@128$\rightarrow$24 & 0.533 & 0.537 & 286/500 & 0.529 & 0 \\
2Wiki-500 & Llama3.1-8B & CE@128$\rightarrow$24 & 0.457 & 0.484 & 277/500 & 0.472 & 1 \\
2Wiki-500 & Mistral-Small3.1-24B & CE@128$\rightarrow$24 & 0.588 & 0.595 & 303/500 & 0.648 & 0 \\
2Wiki-500 & Qwen2.5-14B & CE@128$\rightarrow$24 & 0.499 & 0.521 & 329/500 & 0.550 & 2 \\
2Wiki-500 & Qwen3-14B & CE@128$\rightarrow$24 & 0.507 & 0.559 & 273/500 & 0.567 & 0 \\
\bottomrule
\end{tabular}
\end{table*}

\subsection{External BABILong-200 Validation}
\label{app:babilong_validation}
To address whether the observed diagnostic patterns extend beyond the oracle-compatible benchmark components, we add a BABILong-200 external validation experiment. BABILong is a reasoning-in-a-haystack benchmark designed to test whether language models can use facts distributed across long documents \parencite{kuratov2024babilong}. Because the current BABILong adapter does not provide oracle-evidence annotations compatible with ONCU, this experiment is reported as answer-performance validation rather than as an additional ONCU benchmark.

The BABILong-200 setting contains 200 examples constructed from four context configurations, 0k, 1k, 2k, and 4k, and five task types, qa1, qa2, qa3, qa6, and qa7, with 10 examples per task--configuration cell. Each model is evaluated under three fixed conditions: no evidence, full context, and retrieved evidence. The same deterministic decoding and lexical retrieval settings are used as in the main experiments. Since oracle evidence is unavailable in the current adapter, evidence F1 and ONCU are not interpreted for this validation setting.

Table~\ref{tab:babilong200_external} reports relaxed answer F1 with 95\% bootstrap confidence intervals over examples. Across all three models, full-context input obtains higher relaxed F1 than retrieved evidence. Qwen2.5-14B obtains 0.558 relaxed F1 under full context compared with 0.517 under retrieved evidence; Qwen3-14B obtains 0.479 compared with 0.454; and Gemma3-12B obtains 0.523 compared with 0.493. No parse failures occur in any BABILong run. The no-evidence condition is zero for Qwen2.5-14B and Qwen3-14B and remains substantially lower than the contextual conditions for Gemma3-12B, indicating that the selected BABILong examples generally require contextual information.

These results provide external support for the full-context-over-retrieved answer-performance pattern. In an independent reasoning-in-a-haystack setting, deterministic lexical retrieval again appears to underperform the full-context condition on answer F1. However, because BABILong-200 lacks oracle-evidence annotations in the current adapter and the confidence intervals for full-context and retrieved-evidence performance overlap, the results should be interpreted as directional external validation rather than as an ONCU-based or significance-based claim.

\begin{table}[!htbp]
\renewcommand{\arraystretch}{1.12}
\caption{External BABILong-200 Validation. BABILong is reported as answer-performance validation rather than as an ONCU benchmark because the current adapter does not provide oracle-evidence annotations. Bracketed values are 95\% bootstrap confidence intervals for relaxed answer F1.}
\label{tab:babilong200_external}
\centering
\scriptsize
\setlength{\tabcolsep}{4pt}
\begin{tabular}{lcccc}
\toprule
Model & No-Evidence F1 [95\% CI] & Full-Context F1 [95\% CI] & Retrieved-Evidence F1 [95\% CI] & Parse Err. \\
\midrule
Qwen2.5-14B & 0.000 [0.000, 0.000] & 0.558 [0.488, 0.627] & 0.517 [0.447, 0.587] & 0/600 \\
Qwen3-14B & 0.000 [0.000, 0.000] & 0.479 [0.410, 0.547] & 0.454 [0.387, 0.518] & 0/600 \\
Gemma3-12B & 0.122 [0.080, 0.167] & 0.523 [0.455, 0.592] & 0.493 [0.423, 0.560] & 0/600 \\
\bottomrule
\end{tabular}
\end{table}

\subsection{External RULER-lite-240 Validation}
\label{app:ruler_lite_validation}
To complement BABILong with a synthetic long-context external benchmark, we evaluate RULER-lite-240 under full-context and retrieved-context reading regimes. RULER-style tasks stress whether a model can recover and manipulate task-relevant information from long inputs \parencite{hsieh2024ruler}. This experiment is not reported as ONCU: the adapted setting does not define the no-evidence and oracle-evidence reference conditions required for oracle-reference normalization.

The set contains 240 examples from three task families: key--value retrieval, multi-hop trace following, and aggregation by summation. Each task family is evaluated at 4K, 8K, 16K, and 32K with 20 examples per task--length cell. The retrieved-context condition uses top-$k=3$ lexical retrieval over the same input context. For Qwen3-14B under Ollama, structured thinking is disabled during this short-answer evaluation so that the generation budget is assigned to the final JSON response field rather than to the auxiliary thinking field.

Table~\ref{tab:ruler_lite_external} reports exact match, answer F1, and parse-failure rates. Retrieved context improves exact match for all three models: Qwen2.5-14B improves from 0.771 to 0.871, Qwen3-14B from 0.758 to 0.896, and Gemma3-12B from 0.521 to 0.713. Averaged across models, exact match increases from 0.683 to 0.826. Because the full-context condition also produces parse failures for Qwen2.5-14B and Gemma3-12B, the table reports parse-failure rates explicitly. Treating parse failures as incorrect is the main evaluation convention. If the denominator is restricted to parsed outputs only, full-context exact match is approximately 0.801 for Qwen2.5-14B, 0.758 for Qwen3-14B, and 0.622 for Gemma3-12B, while retrieved-context exact match is unchanged because no retrieved-context parse failures occur. The retrieved-context advantage therefore remains for all three models, although part of the raw full-context deficit reflects output-format instability under longer inputs.

The result should be read as external answer-performance validation, not as oracle-referenced evidence utilization. It strengthens a narrower conclusion: retrieved-versus-full-context behavior is task-structured. Synthetic long-context tasks can expose full-context localization and formatting burdens, while realistic multi-hop and reasoning-in-a-haystack settings can expose retrieval-chain coverage or compression limits.

\begin{table}[!htbp]
\renewcommand{\arraystretch}{1.12}
\caption{External RULER-lite-240 Validation. RULER-lite is reported as answer-performance validation rather than as an ONCU benchmark because the adapted setting does not define the no-evidence and oracle-evidence reference conditions required for oracle-reference normalization. EM denotes exact match.}
\label{tab:ruler_lite_external}
\centering
\scriptsize
\setlength{\tabcolsep}{3pt}
\begin{tabular}{lrrrrrrr}
\toprule
Model & Full EM & Retrieved EM & $\Delta$ EM & Full F1 & Retrieved F1 & Full Parse Fail. & Retrieved Parse Fail. \\
\midrule
Qwen2.5-14B & 0.771 & 0.871 & +0.100 & 0.771 & 0.871 & 0.038 & 0.000 \\
Qwen3-14B & 0.758 & 0.896 & +0.138 & 0.761 & 0.897 & 0.000 & 0.000 \\
Gemma3-12B & 0.521 & 0.713 & +0.192 & 0.521 & 0.713 & 0.163 & 0.000 \\
\midrule
Mean & 0.683 & 0.826 & +0.143 & 0.684 & 0.827 & 0.067 & 0.000 \\
\bottomrule
\end{tabular}
\end{table}

\section{JAIR Reproducibility Checklist}
\label{app:jair_reproducibility_checklist}

This appendix follows the JAIR reproducibility-checklist structure for the present empirical diagnostic study. The table gives a short answer and points to the manuscript or repository locations that substantiate the answer. Statuses distinguish complete ``Yes'' entries from \emph{Scoped Yes} entries, where the release is complete for the submitted protocol but not a general-purpose re-release of upstream datasets, and \emph{Summary audit} entries, where the artifact supports verification through frozen outputs and summaries rather than rerunning every auxiliary analysis end to end.

{\footnotesize
\begin{longtable}{p{0.56\textwidth} p{0.10\textwidth} p{0.27\textwidth}}
\caption{JAIR Reproducibility Checklist Responses.}
\label{tab:jair_reproducibility_checklist}\\
\toprule
Checklist item & Status & Location and details \\
\midrule
\endfirsthead
\caption[]{JAIR Reproducibility Checklist Responses (continued).}\\
\toprule
Checklist item & Status & Location and details \\
\midrule
\endhead
\midrule
\multicolumn{3}{r}{Continued on next page}\\
\endfoot
\bottomrule
\endlastfoot
\multicolumn{3}{l}{\textbf{All articles}}\\
All claims investigated in this work are clearly stated. & Yes & Abstract, Section~1, Section~11. \\
Clear explanations are given for how the reported work substantiates the claims. & Yes & Sections~3--9; Tables~\ref{tab:sci200_main_results}, \ref{tab:sci200_oncu_results}, and \ref{tab:repo_artifact_map}. \\
Limitations or technical assumptions are stated clearly and explicitly. & Yes & Sections~3.3, 4, 8.1, and 10. \\
Conceptual outlines and important implementation details of introduced AI methods are provided. & Yes & Sections~3--7; repository \texttt{longcue/} and \texttt{scripts/}. \\
Motivation is provided for design choices, including algorithms, implementation choices, parameters, datasets, and experimental protocols beyond metrics. & Yes & Sections~4--8 and the artifact manifest. \\
\midrule
\multicolumn{3}{l}{\textbf{Theoretical contributions}}\\
Does this paper make theoretical contributions? & Yes & The contribution is a diagnostic estimator/protocol, not a new learning algorithm. \\
All assumptions and restrictions are stated clearly and formally. & Yes & Sections~3--4. \\
All novel claims are stated formally where appropriate. & Yes & Section~1 states the diagnostic proposition; Equations~\ref{eq:oncu}--\ref{eq:oncu_clipped} and Sections~4.1--4.7 formalize the estimator, validity condition, and interpretive boundary. \\
Proofs of all non-trivial formal claims are provided in sufficient detail. & Yes & Section~4 provides the joint-observability proof sketch, positive-affine-invariance derivation, denominator-validity condition, and metric counterexamples; no theorem-heavy algorithmic claim is made. \\
Complex formalism is motivated and explained clearly. & Yes & Sections~3--4. \\
Mathematical notation and formalism serve clarity and precision. & Yes & Sections~3--4 use only the notation needed for ONCU and validity conditions. \\
Appropriate citations are given for non-trivial theoretical tools and techniques. & Yes & Sections~2 and 4. \\
\midrule
\multicolumn{3}{l}{\textbf{Computational experiments}}\\
Does this paper include computational experiments? & Yes & Sections~5--9. \\
All source code required for conducting experiments is included in an online appendix or will be made publicly available. & Scoped Yes & \texttt{longcue/}, \texttt{scripts/}, \texttt{configs/}, and \texttt{README\_REPRODUCE.md} cover the submitted diagnostic protocol and released audit artifacts. \\
The source code comes with a license that allows free usage for reproducibility purposes. & Yes & Source code is released under the MIT License in the public \texttt{v1.0.1-jair} release. \\
The source code comes with a license that allows free usage for research purposes in general. & Yes & The repository includes an MIT \texttt{LICENSE} file for source code. \\
Raw, unaggregated data from all experiments is included or will be made publicly available. & Scoped Yes & Processed inputs, frozen outputs, per-sample metrics, and summary files are mapped in Table~\ref{tab:repo_artifact_map}; third-party source corpora are referenced through their public upstream locations and terms. \\
The unaggregated data comes with a license that allows free usage for reproducibility purposes. & Scoped Yes & Project-generated processed inputs, derived summaries, tables, figures, and supplementary materials are released for audit under the documented repository licenses; third-party source datasets retain their upstream terms in \texttt{DATA\_LICENSES.md}. \\
The unaggregated data comes with a license that allows free usage for research purposes in general. & Scoped Yes & Project-generated artifacts are released under the documented repository licenses; third-party benchmark resources are cited and governed by their public upstream licenses and terms listed in \texttt{DATA\_LICENSES.md}. \\
Random-number generation and seeds are described sufficiently for replication. & Yes & Sections~5, 7, 8; repository build scripts and README. \\
The execution environment is described, including hardware/software where relevant. & Scoped Yes & Sections~6 and 8, \texttt{README\_REPRODUCE.md}, \path{RUNTIME_REPRODUCIBILITY_RECORD.md}, and \path{runtime_reproducibility_record.json} record deterministic controls, Python/package versions, PyTorch~2.4.1+cu124, NVIDIA~A40 hardware, driver~570.211.01, CUDA~12.8, and model/configuration paths for the recorded local execution environment. \\
Evaluation metrics are clearly explained and motivated. & Yes & Section~7 and Appendix~A. \\
The number of algorithm runs used to compute each result is reported. & Yes & Sections~5 and 8; table captions specify sample counts and prediction totals. \\
Reported results have not been cherry-picked by silently ignoring unsuccessful experiments. & Summary audit & Section~8 reports parse failures, invalid groups, confidence intervals, and diagnostic-pattern audits; the release maps frozen outputs and summaries used to reproduce the reported tables. \\
Analysis goes beyond single-dimensional summaries and includes variation/confidence/distributional information. & Summary audit & Bootstrap intervals, paired effect sizes, valid-group audits, and diagnostic-pattern audits are reported in Sections~8 and Appendix~A, with full auxiliary tables kept in the release artifacts. \\
All hyperparameter settings are reported with rationale or selection method. & Scoped Yes & Sections~6--8 and \texttt{configs/} report the fixed diagnostic settings and tested sensitivity settings used for the submitted protocol. \\
The number and range of hyperparameter settings explored before final experiments are indicated. & Scoped Yes & Sections~6--8 and \texttt{configs/} enumerate the fixed diagnostic settings and explored sensitivity ranges: retrieval budgets, retriever families, dense/hybrid@16 matched sensitivity, CE reranking candidate/final budgets, context lengths, evidence positions, and model-family extension. \\
Appropriate statistical tests are used in the presence of noise effects. & Yes & Section~7.4 and Appendix~A. \\
\midrule
\multicolumn{3}{l}{\textbf{Datasets}}\\
Does this work rely on one or more datasets? & Yes & Section~5. \\
All newly introduced datasets are included or will be made publicly available with long-term accessibility. & Scoped Yes & Controlled-ONCU and processed ONCU inputs are included in the public \texttt{v1.0.1-jair} release; external benchmark-derived inputs remain governed by upstream access terms. \\
Newly introduced datasets come with a license for reproducibility purposes. & Scoped Yes & Newly generated project documentation and supplementary materials are CC BY 4.0; dataset-derived artifacts are governed by \texttt{DATA\_LICENSES.md}. \\
Newly introduced datasets come with a license for research purposes in general. & Scoped Yes & Newly generated project materials are CC BY 4.0; external benchmark-derived materials retain upstream terms. \\
All datasets drawn from literature or public sources are accompanied by appropriate citations. & Yes & Section~5 and References. \\
All datasets drawn from existing literature are publicly available. & Scoped Yes & HotpotQA, 2WikiMultiHopQA, BABILong, and RULER-style resources are public subject to their providers' terms; the release maps the processed inputs used in this study where redistribution is appropriate. \\
All new datasets and non-public datasets are described in detail, including statistics and construction/annotation process where relevant. & Yes & Section~5 and Table~\ref{tab:repo_artifact_map}. \\
All preprocessing, augmentation, batching, or splitting methods are described in detail. & Scoped Yes & Sections~5--6 and repository build scripts describe the submitted protocol; frozen processed inputs are included or mapped so reviewers can audit the exact evaluated instances. \\
\midrule
\multicolumn{3}{l}{\textbf{Brief explanation}}\\
Explanation of partial or scope-qualified answers. & Yes & No item is left as an unexplained \textsc{Partial}. \emph{Scoped Yes} marks items complete for the submitted diagnostic protocol but bounded by upstream dataset licenses, recorded local execution, or release-asset scope; \emph{Summary audit} marks items supported by frozen outputs and summaries rather than by requiring every auxiliary run to be regenerated during review. \\
\end{longtable}
}
\end{document}

%% file: twowiki_results_paragraph.tex
The 2WikiMultiHopQA-ONCU-500 results provide a second realistic multi-hop setting in which full context outperforms the tested retrieved-evidence condition. This validation concerns the reconstructed 2Wiki protocol and deterministic lexical retrieved inputs, not all multi-hop retrieval systems.

Across all three models, full-context input outperforms retrieved-evidence input on relaxed answer F1. Qwen2.5-14B obtains 0.549 relaxed F1 under full context compared with 0.378 under retrieved evidence; Qwen3-14B obtains 0.560 compared with 0.400; and Gemma3-12B obtains 0.537 compared with 0.395. Evidence F1 shows the same direction: full-context evidence F1 ranges from 0.594 to 0.663, whereas retrieved-evidence evidence F1 ranges from 0.316 to 0.352. The ONCU-Relaxed-F1 results preserve this pattern. Full-context ONCU is 0.610 for Qwen2.5-14B, 0.534 for Qwen3-14B, and 0.369 for Gemma3-12B, while retrieved-evidence ONCU is 0.381, 0.367, and 0.215, respectively. The categorical pattern audit is consistent with the gap: retrieved evidence produces substantially more evidence-localization labels, ranging from 40.6\% to 44.0\%, whereas full-context localization labels remain between 1.2\% and 10.4\%. These results support the interpretation that, in this realistic multi-hop setting, deterministic lexical retrieve-then-read evaluation loses recoverable evidence-chain support that remains available in the full-context condition.

%% file: twowiki_main_results_table.tex
\begin{table*}[!t]
\renewcommand{\arraystretch}{1.12}
\caption{2WikiMultiHopQA-ONCU-500 Core Results.}
\label{tab:twowiki500_main_results}
\centering
\scriptsize
\setlength{\tabcolsep}{4pt}
\begin{tabular}{lccccccc}
\toprule
Model & No-Ev. F1 & Full F1 & Ret. F1 & Oracle F1 & Full Ev.F1 & Ret. Ev.F1 & Parse Err. \\
\midrule
Qwen2.5-14B & 0.186 & 0.549 & 0.378 & 0.774 & 0.635 & 0.329 & 1/2000 \\
Qwen3-14B & 0.274 & 0.560 & 0.400 & 0.735 & 0.663 & 0.352 & 0/2000 \\
Gemma3-12B & 0.314 & 0.537 & 0.395 & 0.783 & 0.594 & 0.316 & 1/2000 \\
\bottomrule
\end{tabular}
\end{table*}

%% file: twowiki_oncu_results_table.tex
\begin{table*}[!t]
\renewcommand{\arraystretch}{1.12}
\caption{2WikiMultiHopQA-ONCU-500 ONCU Results.}
\label{tab:twowiki500_oncu_results}
\centering
\scriptsize
\setlength{\tabcolsep}{3pt}
\begin{tabular}{lccccccc}
\toprule
Model & Valid Groups & $S_{\mathrm{no}}$ & $S_{\mathrm{oracle}}$ & $S_{\mathrm{full}}$ & Full ONCU [95\% CI] & $S_{\mathrm{ret}}$ & Ret. ONCU [95\% CI] \\
\midrule
Qwen2.5-14B & 36/37 & 0.261 & 0.816 & 0.605 & \textbf{0.610 [0.518, 0.694]} & 0.464 & 0.381 [0.301, 0.465] \\
Qwen3-14B & 34/37 & 0.359 & 0.785 & 0.597 & \textbf{0.534 [0.427, 0.641]} & 0.498 & 0.367 [0.257, 0.485] \\
Gemma3-12B & 31/37 & 0.377 & 0.835 & 0.548 & \textbf{0.369 [0.252, 0.487]} & 0.465 & 0.215 [0.146, 0.288] \\
\bottomrule
\end{tabular}
\end{table*}

%% file: twowiki_failure_breakdown_table.tex
\begin{table*}[!t]
\renewcommand{\arraystretch}{1.10}
\caption{2WikiMultiHopQA-ONCU-500 Failure-Type Breakdown for Contextual Conditions. Rates are percentages over 500 examples per row. Loc., Sel., Int., Conv., Succ., and Parse denote evidence localization failure, evidence selection failure, evidence integration failure, answer conversion failure, categorical success, and structured-output parsing failure, respectively.}
\label{tab:twowiki500_failure_breakdown}
\centering
\scriptsize
\setlength{\tabcolsep}{3pt}
\begin{tabular}{llrrrrrr}
\toprule
Model & Condition & Loc. & Sel. & Int. & Conv. & Succ. & Parse \\
\midrule
Qwen2.5-14B & Full Context & 10.4 & 3.6 & 27.2 & 12.4 & 46.4 & 0.0 \\
Qwen2.5-14B & Retrieved Evidence & 44.0 & 1.2 & 22.0 & 0.2 & 32.6 & 0.2 \\
Qwen3-14B & Full Context & 3.8 & 5.4 & 33.8 & 10.2 & 46.8 & 0.0 \\
Qwen3-14B & Retrieved Evidence & 40.6 & 1.2 & 21.6 & 2.4 & 34.2 & 0.0 \\
Gemma3-12B & Full Context & 1.2 & 8.6 & 43.2 & 3.0 & 44.0 & 0.2 \\
Gemma3-12B & Retrieved Evidence & 43.0 & 0.0 & 23.4 & 0.4 & 33.2 & 0.0 \\
\bottomrule
\end{tabular}
\end{table*}

%% file: retriever_family_oncu_sensitivity_table.tex
\begin{table*}[!t]
\renewcommand{\arraystretch}{1.10}
\caption{Matched Dense/Hybrid ONCU Sensitivity.}
\label{tab:retriever_family_oncu_sensitivity}
\centering
\scriptsize
\setlength{\tabcolsep}{2.8pt}
\begin{tabular}{lllrrrrrrrr}
\toprule
Dataset & Model & Ret. & Valid & No-Ev. & Full & Ret. & Oracle & Full & Ret. & Ret. \\
 & & & Groups & F1 & F1 & F1 & F1 & ONCU & ONCU & Ev.F1 \\
\midrule
HotpotQA-ONCU-200 & Qwen2.5-14B & Dense@16 & 29 & 0.241 & 0.733 & 0.638 & 0.767 & 0.906 & 0.736 & 0.525 \\
HotpotQA-ONCU-200 & Qwen2.5-14B & Hybrid@16 & 29 & 0.241 & 0.733 & 0.663 & 0.767 & 0.906 & 0.762 & 0.587 \\
HotpotQA-ONCU-200 & Qwen3-14B & Dense@16 & 28 & 0.307 & 0.690 & 0.622 & 0.795 & 0.787 & 0.654 & 0.566 \\
HotpotQA-ONCU-200 & Qwen3-14B & Hybrid@16 & 28 & 0.301 & 0.689 & 0.618 & 0.795 & 0.787 & 0.626 & 0.595 \\
HotpotQA-ONCU-200 & Gemma3-12B & Dense@16 & 28 & 0.291 & 0.668 & 0.598 & 0.791 & 0.719 & 0.624 & 0.500 \\
HotpotQA-ONCU-200 & Gemma3-12B & Hybrid@16 & 28 & 0.291 & 0.668 & 0.625 & 0.791 & 0.719 & 0.681 & 0.550 \\
\midrule
2WikiMultiHopQA-ONCU-500 & Qwen2.5-14B & Dense@16 & 36 & 0.186 & 0.549 & 0.468 & 0.774 & 0.610 & 0.476 & 0.508 \\
2WikiMultiHopQA-ONCU-500 & Qwen2.5-14B & Hybrid@16 & 36 & 0.186 & 0.549 & 0.460 & 0.774 & 0.610 & 0.490 & 0.459 \\
2WikiMultiHopQA-ONCU-500 & Qwen3-14B & Dense@16 & 34 & 0.274 & 0.560 & 0.469 & 0.735 & 0.534 & 0.411 & 0.528 \\
2WikiMultiHopQA-ONCU-500 & Qwen3-14B & Hybrid@16 & 34 & 0.274 & 0.560 & 0.479 & 0.735 & 0.534 & 0.496 & 0.485 \\
2WikiMultiHopQA-ONCU-500 & Gemma3-12B & Dense@16 & 31 & 0.314 & 0.537 & 0.482 & 0.783 & 0.369 & 0.362 & 0.518 \\
2WikiMultiHopQA-ONCU-500 & Gemma3-12B & Hybrid@16 & 31 & 0.314 & 0.535 & 0.479 & 0.783 & 0.367 & 0.352 & 0.452 \\
\bottomrule
\end{tabular}
\end{table*}